\documentclass{article}

\usepackage[nonatbib,final]{nips_2016}
\usepackage[square,numbers,sort&compress]{natbib}

\usepackage{mathrsfs}
\usepackage{amssymb}
\usepackage{bbm}
\usepackage{graphicx}
\usepackage[ruled,vlined]{algorithm2e}

\usepackage{amsmath}
\usepackage{graphicx}
\usepackage{dcolumn}
\usepackage{bm}

\newcommand{\brc}[1]{\left<#1\right>}

\newcommand{\fw}{0.6}
\newcommand{\hfw}{0.4}
\newcommand{\tfw}{0.32}

\newcommand{\be}{\begin{equation}}
\newcommand{\ee}{\end{equation}}
\newcommand{\bea}{\begin{eqnarray}}
\newcommand{\eea}{\end{eqnarray}}

\newcommand{\cin}{c_ {\textrm{in}}}
\newcommand{\cout}{c_ {\textrm{out}}}
\newcommand{\pin}{p_{\textrm{in}}}
\newcommand{\pout}{p_{\textrm{out}}}

\newcommand{\argmax}{\arg\!\max}

\newcommand{\myvspace}[1]{}

\title{Robust Spectral Detection of Global Structures in the Data
by Learning a Regularization}
\author{
Pan Zhang\\
Institute of Theoretical Physics, Chinese Academy of Sciences, Beijing 100190, China\\
  \texttt{panzhang@itp.ac.cn} \\
}
\begin{document}
\maketitle
\begin{abstract}
Spectral methods are popular in detecting global structures in the given data that can be represented as a matrix.
However when the data matrix is sparse or noisy,
classic spectral methods usually fail to work, due to localization of eigenvectors (or singular vectors)
induced by the sparsity or noise.
In this work, we propose
a general method to solve the localization problem by learning a regularization matrix from the localized eigenvectors. 
Using matrix perturbation analysis, we demonstrate that the learned 
regularizations suppress down the eigenvalues associated with localized 
eigenvectors and enable us to recover the informative eigenvectors representing the global structure.
We show applications of our
method in several inference problems: community detection in networks, clustering from pairwise similarities,
rank estimation and matrix completion problems.
Using extensive experiments, we illustrate that
our method solves the localization problem and works down to the 
theoretical detectability limits
in different kinds of synthetic data. This is
in contrast with existing spectral algorithms based on
data matrix, non-backtracking matrix, Laplacians and those with 
rank-one regularizations, which perform poorly in the sparse 
case with noise.
\end{abstract}
\maketitle
\section{Introduction}
In many statistical inference 
problems, the task is to detect, from given data, a global structure 
such as low-rank structure or clustering.
The task is usually hard to solve since modern datasets usually
have a large dimensionality. When the dataset can be represented 
as a matrix, spectral methods
are popular as it gives a natural way to 
reduce the dimensionality of data using eigenvectors or singular vectors.
In the point-of-view of inference, data can be seen as measurements to the 
underlying structure. Thus more data gives more precise information about
the underlying structure. 

\begin{figure}[h]
   \centering
   \label{fig:spectrum}
    \includegraphics[width=\hfw\linewidth]{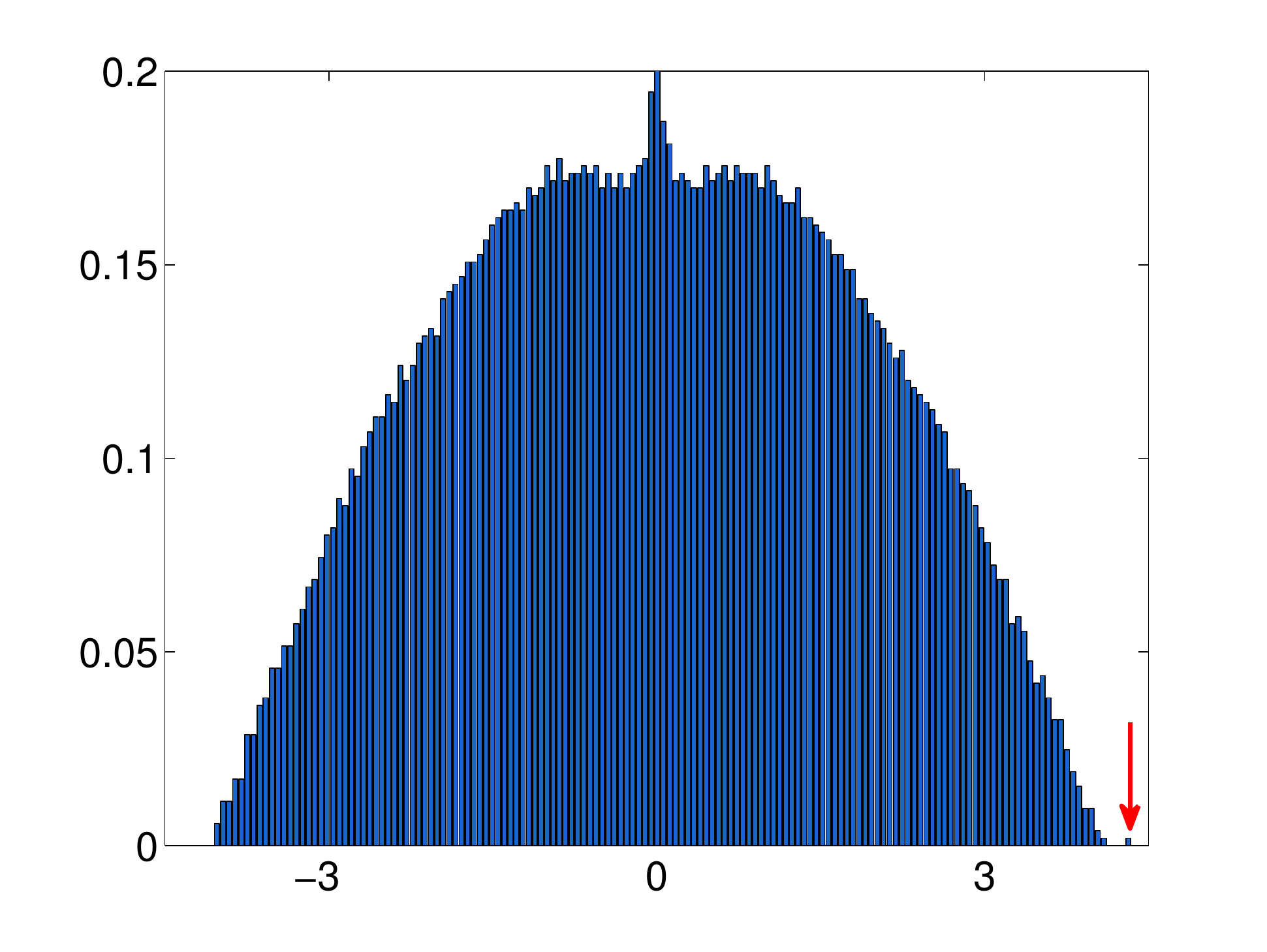} 
	\includegraphics[width=\hfw\linewidth]{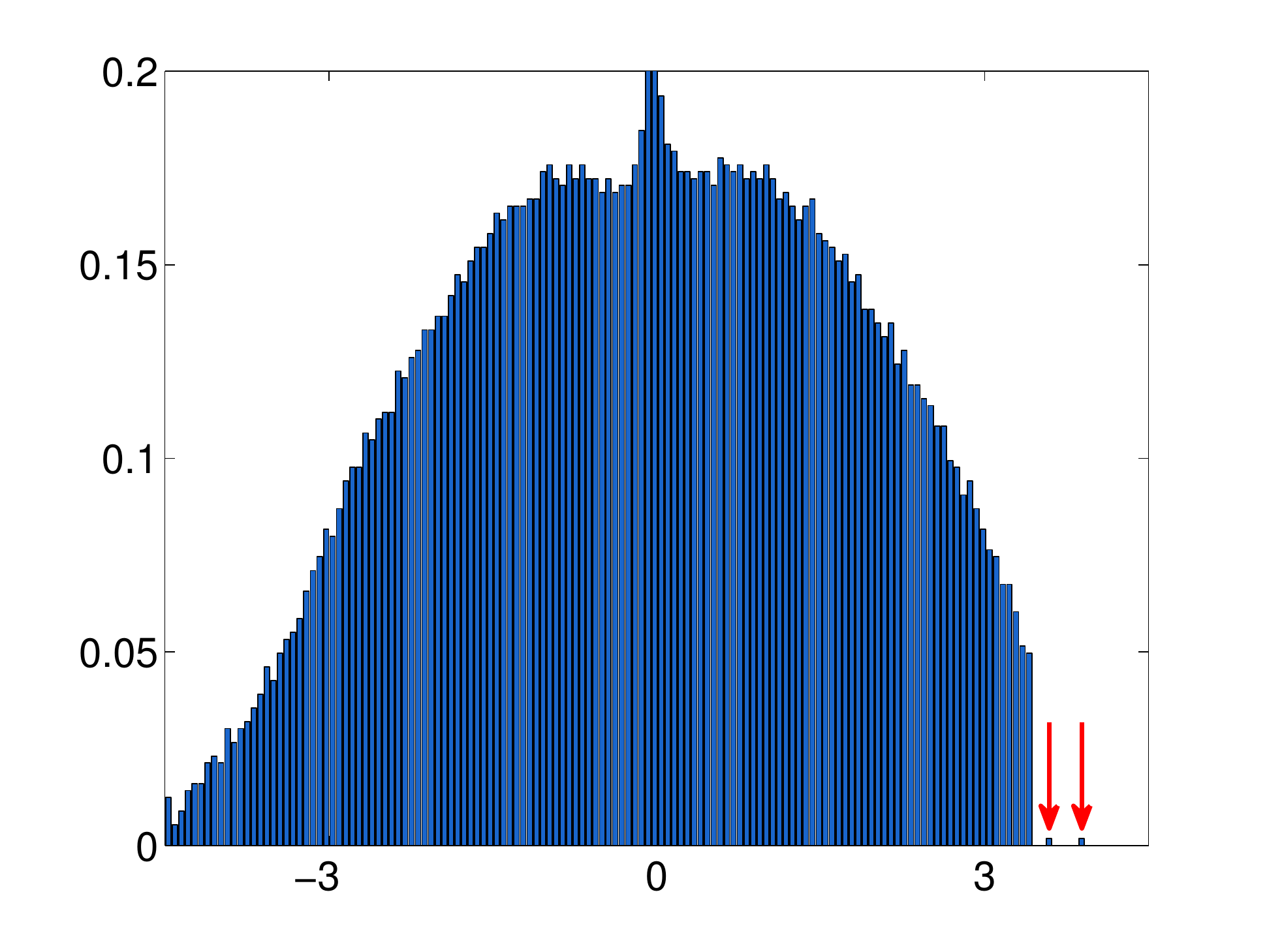} 
\caption{Spectral density of the adjacency matrix (\textit{left}) and X-Laplacian 
(\textit{right}) of a graph generated by the stochastic block model with $n=10000$ nodes, average degree $c=3$, $q=2$ groups and $\epsilon=0.125$. 
Red arrows point to eigenvalues out of the bulk.
}
\end{figure}

However in many situations when we do not have enough
measurements, i.e. the data matrix is sparse, 
standard spectral methods usually have localization problems thus do not work well.
One example is the community detection in sparse networks, where
the task is to partition nodes 
into groups such that there are many edges connecting nodes within 
the same group and comparatively few edges connecting nodes in
different groups.
It is well known that when the graph has a large connectivity $c$, 
simply using the first 
few eigenvectors of the adjacency matrix $A\in \{0,1\}^{n\times n}$ 
(with $A_{ij}=1$ denoting an edge between node $i$ and node $j$,and $A_{ij}=0$ 
otherwise) gives a good result. 
In this case, like that of a sufficiently dense Erd\H{o}s-R\'enyi (ER)
random graph with average degree $c$, the spectral density follows Wigner's semicircle rule,
$P(\lambda)=\sqrt{4c-\lambda^2}/2\pi c$, 
and there is a gap between the edge of bulk of eigenvalues and 
the informative eigenvalue that represents the underlying community structure.
However
when the network is large and sparse,
the spectral density of the adjacency matrix deviates from the
semicircle, the informative eigenvalue is hidden in the bulk of eigenvalues, as displayed in Fig.~\ref{fig:spectrum} left. Its 
eigenvectors associated with largest eigenvalues (which are roughly 
proportional to $\log n/\log\log n$ for ER random graphs) are localized on the large-degree nodes, thus
reveal only local structures about large degrees rather than the underlying
global structure.
Other standard matrices for spectral clustering \cite{Luxburg2007,ng2002spectral},
e.g. Laplacian, random walk matrix, normalized Laplacian, all have localization problems
but on different local structures such as dangling trees.

Another example is the matrix completion problem
which asks to infer missing entries of matrix 
$A\in\mathbb R^{m\times n}$ with rank $r\ll \sqrt{mn}$ from only few observed entries. 
A popular method for this problem is based on
the singular value decomposition (SVD) of the data matrix. However it is well known that
when the matrix is sparse, SVD-based method performs very poorly, because the
singular vectors corresponding to the largest singular values 
are localized, i.e. highly concentrated on high-weight column or row indices.

A simple way to ease the pain of localization induced by high 
degree or weight is trimming \cite{CPC7183040,keshavan2009low}
which sets to zero columns or rows with a large degree or weight.
However trimming throws away part of the information, thus does not work all the way down to the theoretical limit in the
community detection problem~\cite{CPC7183040,Krzakala2013}. It also performs
worse than other methods in matrix completion problem~\cite{saade2015matrix}.

In recent years, many methods have been proposed for the sparsity-problem.
One kind of methods use new
linear operators related to the belief propagation and Bethe free energy, such as the non-backtracking matrix \cite{Krzakala2013} 
and Bethe Hessian \cite{saade2014spectral}. Another kind of 
methods add to the data matrix or its variance
a rank-one regularization matrix
~\cite{amini2013pseudo,le2015sparse,joseph2013impact,le2015concentration,lei2014consistency,qin2013regularized}.
These methods are quite successful in some inference problems in
the sparse regime.
However in our understanding none of them works
in a general way to solve the localization problem.
For instance, the non-backtracking matrix and the Bethe Hessian 
work very well when
the graph has a locally-tree-like structure, but they have again the localization
problems when the system has short loops or sub-structures like triangles and cliques.
Moreover its performance is sensitive to the noise in the 
data~\cite{Javanmard21032016}. Rank-one regularizations
have been used for a long time in practice, the most famous example is 
the ``teleportation'' term in the Google matrix. However there is no 
satisfactory way to determine the optimal amount of regularization in general.
Moreover, analogous to the non-backtracking matrix and Bethe Hessian, the rank-one regularization approach is
also sensitive to the noise, as we will show in the paper.

The main contribution of this paper is to illustrate how to solve the 
localization problem of spectral methods for general inference problems in
sparse regime and with noise, by learning a proper regularization that
is specific for the given data matrix from its localized eigenvectors.
%
In the following text we will first discuss in Sec.~\ref{sec:reg} that
all three methods for community detection in sparse graphs can 
be put into the framework of regularization.
Thus the drawbacks of existing methods can be seen as improper choices
of regularizations.
In Sec.~\ref{sec:our} we investigate how to
choose a good regularization that is dedicated for the given data, 
rather than taking a fixed-form regularization as in the 
existing approaches.
We use matrix perturbation analysis to illustrate how the regularization
works in penalizing the localized eigenvectors, and making 
the informative eigenvectors that correlate with the global structure
float to the top positions in spectrum.
In Sec.~\ref{sec:num} we use extensive numerical experiments to 
validate our approach on
several well-studied inference problems, including the community detection in 
sparse graphs, clustering from sparse pairwise entries, rank estimation
and matrix completion from few entries. 

\section{Regularization as a unified framework}
\label{sec:reg}
We see that the above three methods for
the community detection problem in sparse graphs,
i.e. trimming, non-backtracking/Bethe Hessian, and rank-one regularizations,
can be understood as doing different ways of regularizations.
In this framework, we consider a regularized matrix 
\begin{equation}
	\label{eq:L}
	L=\hat A+\hat R.
\end{equation} 
Here matrix $\hat A$ is the data matrix or its (symmetric) variance, 
such as $\tilde A=D^{-1/2}AD^{-1/2}$ with $D$ denoting the diagonal matrix
of degrees, and matrix $\hat R$ is a regularization matrix.
The rank-one regularization approaches
~\cite{amini2013pseudo,le2015sparse,joseph2013impact,le2015concentration,lei2014consistency,qin2013regularized} fall naturally into this framework as they
set
$R$ to be a rank-one matrix, $-\zeta \bm 1\bm {1^T}$, with $\zeta$ being
a tunable parameter controlling strength of regularizations. It is also
easy to see that in the trimming, $\hat A$ is set to be 
the adjacency matrix and $\hat R$ contains entries to 
remove columns or rows with high degrees from $A$.

For spectral algorithms using the non-backtracking matrix,
its relation to form Eq.~\eqref{eq:L} is not straightforward.
However we can link them using the theory of graph zeta 
function~\cite{hashimoto1989zeta} which says that an eigenvalue $\mu$ of
the non-backtracking operator satisfies the following quadratic eigenvalue equation,
$$\mathrm{det}[\mu^2 I-\mu A+(D-I)]=0,$$
where $I$ is the identity matrix.
It indicates that a particular vector $v$ that is related to
the eigenvector of the non-backtracking matrix satisfies
$(A-\frac{D-I}{\mu})v=\mu v$.
Thus spectral clustering algorithm using the non-backtracking
matrix is equivalent to the spectral clustering algorithm using matrix 
with form in Eq.~\eqref{eq:L}, while $\hat A=A$,
$\hat R=\frac{D-I}{\mu}$, and $\mu$ acting as a parameter.
We note here that the parameter does not necessarily be an eigenevalue of
the non-backtracking matrix. Actually a range of parameters work well in practice, like those estimated from the spin-glass transition of the system
\cite{saade2014spectral}.
So we have related different approaches of resolving localizations
of spectral algorithm in sparse graphs into the framework of regularization.
Although this relation is in the context of community detection
in networks, we think it is a general point-of-view, when
the data matrix has a general form
rather than a $\{0,1\}$ matrix.

As we have argued in the introduction, above three ways of regularization
work from case to case and have different problems,
especially when system has noise. It means that in the framework of 
regularizations, the effective regularization matrix $\hat R$ added 
by these methods do not work in a general way and is not robust.
In our understanding, the problem arises from the fact that in all these
methods, the form of regularization is \textit{fixed} for all kinds of
data,
regardless of different reasons for the localization.
Thus one way to solve the problem would be looking for the regularizations
that are specific for the given data, as a feature.
In the following section we will introduce our method
explicitly addressing how to learn such regularizations from localized
eigenvectors of the data matrix.
\section{Learning regularizations from localized eigenvectors}
\label{sec:our}
The reason that the informative eigenvectors are hidden in the bulk is that 
some random eigenvectors have large eigenvalues, due to the localization which
represent the local structures of the system.
In the complementary side, if these eigenvectors are not localized, 
they are supposed to have smaller eigenvalues than the informative ones which reveal the 
global structures of the graph. This is the main assumption that our idea is
based on.

In this work we use the \textit{Inverse Participation Ratio} (IPR), $I(v)=\sum_{i=1}^nv_i^4$, to quantify the amount
of localization of a (normalized) eigenvector $v$. IPR has been used frequently in physics, for example for distinguishing the extended state from the localized state when applied on the wave function~\cite{bell1970atomic}.
It is easy to check that $I(v)$ ranges from $\frac{1}{n}$ for vector 
$ \{\frac{1}{\sqrt{n}},\frac{1}{\sqrt{n}},...,\frac{1}{\sqrt{n}}\}$ to $1$ for vector $\{0,...,0,1,0,...,0\}$. That is, a larger $I(v)$ indicates more localization in vector $v$.

Our idea is to create a matrix $L_X$ with similar structures to $A$, but 
with non-localized leading eigenvectors.
We call the resulting matrix \textit{X-Laplacian}, and define it as $L_X=A+X$,
where matrix $A$ is the data matrix (or its variant), and $X$ is learned 
using the procedure detailed below:

\begin{algorithm}[H]
\caption{Regularization Learning} 
	\KwIn{Real symmetric matrix $A$, number of eigenvectors $q$, learning rate $\eta= O(1)$, threshold $\Delta$. }
	 \KwOut{X-Laplacian, $L_X$, whose leading eigenvectors reveal the global structures in $A$.}
\begin{enumerate}
	\item Set $X$ to be all-zero matrix.
	\item Find set of eigenvectors $U=\{u_1,u_2,...,u_{q}\}$ associated with the first $q$ largest eigenvalues (in algebra) of $L_X$.
	\item Identify the eigenvector $v$ that has the largest inverse participation ratio among the $q$ eigenvectors in $U$. That is, find $v=\argmax_{u\in U}I(u).$
	\item if $I(v)<\Delta$, return $L_X=A+X$; Otherwise, $\forall i, X_{ii}\leftarrow X_{ii}-\eta v_i^2$, then go to step 2.
\end{enumerate}
\end{algorithm}
We can see that the regularization matrix $X$ is a diagonal matrix, its diagonal
entries are learned gradually from the most localized vector among the 
first several eigenvectors.
The effect of $X$ is to penalize the localized eigenvectors, 
by suppressing down the eigenvalues associated with the localized eigenvectors. The learning will continue until all $q$
leading eigenvectors are delocalized, thus are supposed to correlate 
with the global structure rather than the local structures.
As an example, we show the effect of $X$ to the spectrum
in Fig.~\ref{fig:spectrum}. In the left panel, we plot
the spectrum of the adjacency matrix (i.e. before
learning $X$) and the X-Laplacian (i.e. after learning $X$) of a 
sparse network generated by the 
stochastic block model with $q=2$ groups. 
For the adjacency matrix in the left panel, localized eigenvectors
have large eigenvalues and contribute a tail to the semicircle, covering the informative
eigenvalue, leaving only one eigenvalue, which corresponds to the 
eigenvector that essentially sorts vertices according to their degree, 
out of the bulk.
The spectral density of X-Laplacian is shown in the right panel of 
Fig.~\ref{fig:spectrum}. We can see that the right corner of the 
continues part of the spectral density appearing in the spectrum of the adjacency matrix ,
is missing here. 
This is because due to the effect of $X$, the eigenvalues that are 
associated with localized
eigenvectors in the adjacency matrix are pushed into the bulk, maintaining
a gap between the edge of bulk and the informative eigenvalue (being pointed by the left red arrow in the figure).

The key procedure of the algorithm is the learning part in
step $4$, which 
updates diagonal terms of matrix $X$ using the most localized eigenvector $v$.
Throughout the paper, by default we use learning rate $\eta=10$ and 
threshold $\Delta=5/n$.
As $\eta=O(1)$ and $v_i^2=O(1/n)$, we can treat the learned
entries in each step, $\hat L$, as a perturbation to matrix $L_X$.
After applying this perturbation, 
we anticipate that an eigenvalue of
$L$ changes from $\lambda_i$ to $\lambda_i + \hat \lambda_i$, and an eigenvector changes from $u_i$ to
$u_i+\hat u_i$. If we assume that matrix $L_X$ is not ill-conditioned, and the
first few eigenvectors that we care about are distinct, then we have
$\hat \lambda_i=u_i^T\hat Lu_i$.
Derivation of the above expression is straightforward, but for the completeness
we put the derivations in the appendices.
In our algorithm, $\hat L$ is a diagonal matrix with entries $\hat L_{ii}=-\eta v_i^2$ with $v$ denoting
the identified eigenvector who has the largest inverse participation ratio, so last equation can be written as
$	\hat \lambda_i=-\eta \sum_kv_k^2u_{ik}^2$.
For the identified vector $v$, we further have
\begin{equation}
	\hat \lambda_v=-\eta \sum_iv_i^4=-\eta I(v).
\end{equation}
It means the eigenvalue of the identified eigenvector with inverse participation
ratio $I(v)$ is decreased by amount $\eta I(v)$. That is, \textit{the more
localized the eigenvector is, the larger penalty on its eigenvalue}.

In addition to the penalty to the localized eigenvalues, We see
that the leading eigenvectors are delocalizing during learning.
We have analyzed the change of eigenvectors after the perturbation
given by the identified vector $v$, and
obtained (see appendices for the derivations) the change of an 
eigenvector $\hat u_i$ as a function of all the other eigenvalues and eigenvectors,
$	\hat u_{i}
	=\sum_{j\neq i}\frac{\sum_ku_{jk}v_k^2u_{ik}}{\lambda_i-\lambda_j}u_j$.
Then the inverse participation ratio of the new vector $u_i+\hat u_i$ can be
written as\\
\begin{align}
I(u_i+\hat u_i)=I(u_i)-4\eta\sum_{l=1}^n\sum_{j\neq i}\frac{u^2_{jl}v_l^2u^4_{il}}{\lambda_i-\lambda_j}
	-4\eta\sum_{l=1}^n\sum_{j\neq i}\sum_{k\neq l}\frac{u_{il}^3v_k^2u_{jk}u_{ik}u_{jl}}{\lambda_i-\lambda_j}.
\label{eq:signal}
\end{align}
As eigenvectors $u_i$ and $u_j$ are orthogonal to each other, the term $4\eta\sum_{l=1}^n\sum_{j\neq i}\frac{u^2_{jl}v_l^2u^4_{il}}{\lambda_i-\lambda_j}$ can be seen as a signal term
and the last term can be seen 
as a cross-talk noise with zero mean. We see that 
the cross-talk noise has a small variance, and empirically its effect can be neglected.
For the leading eigenvector corresponding to the largest eigenvalue $\lambda_i=\lambda_1$, 
it is straightforward to see that the signal term is strictly positive. 
Thus if the learning is slow enough, 
the perturbation will always decrease the inverse participation
ratio of the leading eigenvector. This is essentially an argument for
convergence of the algorithm.
For other top eigenvectors, i.e. the second and third eigenvectors and so on, though
$\lambda_i-\lambda_j$ is not strictly positive, there are much more positive terms
than negative terms in the sum, thus the signal should be positive with
a high probability.
Thus one can conclude that the process of learning $X$ makes
first few eigenvectors de-localizing.

\begin{figure}[h]
   \centering
    \includegraphics[width=0.3\linewidth]{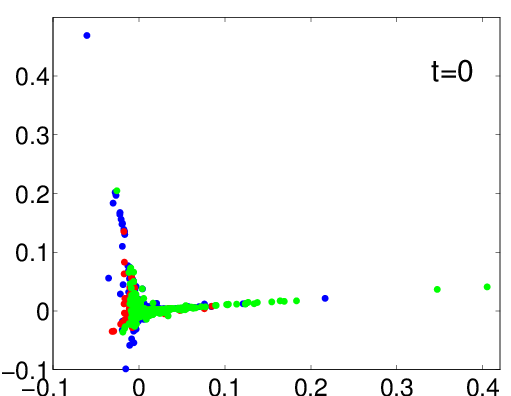} 
    \includegraphics[width=0.31\linewidth]{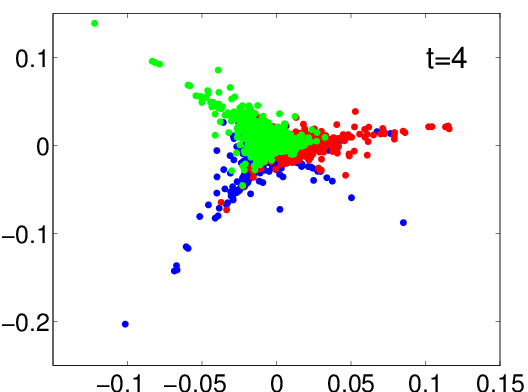} 
    \includegraphics[width=0.3\linewidth]{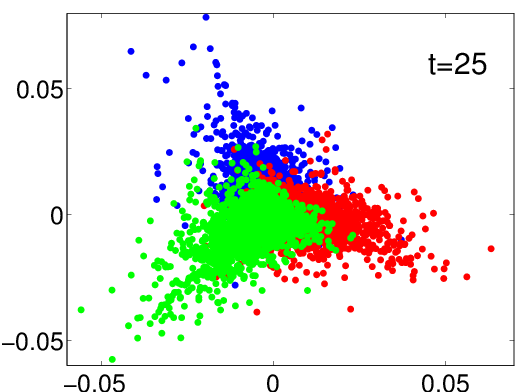} 
	\caption{
   \label{fig:v2v3}
	The second eigenvector $V_2$ compared with the third eigenvector $V_3$ 
	of $L_X$ for a network at three steps with $t=0,4$ and $25$ during learning.
	The network has
$n=42000$ nodes, $q=3$ groups, average degree $c=3$, $\epsilon=0.08$,
three colors represent group labels in the planted partition.
}
\end{figure}

An example illustrating the process of the learning is shown in 
Fig.~\ref{fig:v2v3} where we plot the second eigenvector vs. 
the third eigenvector, at several times steps
during the learning, for a network generated by the stochastic
block model with $q=3$ groups.
We see that at $t=0$, i.e. without learning, both eigenvectors are localized, with a large range of distribution in entries. 
The color of eigenvectors encodes the group membership in
the planted partition. We see that at $t=0$ three colors are mixed together
indicating that two eigenvectors are not correlated with the planted partition.
At $t=4$ 
three colors begin to separate, and range of entry distribution
become smaller, indicating that the localization is lighter. At $t=25$, three colors are more
separated, the partition obtained by applying k-means algorithm using these vectors
successfully recovers $70\%$ of the group memberships. Moreover we can
see that the range of entries of eigenvectors shrink to $[-0.06,0.06]$, 
giving a small inverse participation ratio.

\section{Numerical evaluations}
\label{sec:num}
In this section we validate our approach with experiments on several inference problems, i.e. community detection
problems, clustering from sparse pairwise entries, rank estimation and 
matrix completion from a few entries. 
We will compare performance of spectral algorithms
using the X-Laplacian with recently proposed state-of-the-art spectral methods
in the sparse regime.
\subsection{Community Detection}
First we use synthetic networks generated by the stochastic block model~\cite{holland1983stochastic},
and its variant with noise~\cite{Javanmard21032016}. 
The standard Stochastic Block Model (SBM), also called the planted partition 
model, is a popular model to generate ensemble of networks with community 
structure. There are $q$ groups of nodes and a planted partition 
$\{t^*_i\}\in\{1,...,q\}$. Edges are generated independently according to
a $q\times q$ matrix $\{p_{ab}\}$. Without loss of generality 
here we discuss the commonly studied case where the $q$ 
groups have equal size and where $\{p_{ab}\}$ has only two distinct entries,
$p_{ab}=\cin/n$ if $a=b$ and $\cout/n$ if $a\neq b$.
Given the average degree of the graph, there is a 
so-called detectability transition
$\epsilon^*=\cout/\cin=(\sqrt{c}-1)/(\sqrt{c}-1+q)$ ~\cite{Decelle2011} , beyond which
point it is not possible to obtain any information about the planted
partition. It is also known spectral algorithms based on the non-backtracking
matrix succeed all the way down to the transition~\cite{Krzakala2013}. This transition was recently established rigorously in the
case of $q=2$~\cite{Mossel2012,massoulie2014community}.
Comparisons of spectral methods using different matrices are shown in Fig.~\ref{fig:community_detection} left.
From the figure we see that the X-Laplacian works as well as the non-backtracking
matrix, down to the detectability transition. While the direct use
of the adjacency matrix, i.e. $L_X$ before learning, does not work well when
$\epsilon$ exceeds about $0.1$.

In the right panel of Fig.~\ref{fig:community_detection},
each network is generated by the stochastic block model with 
the same parameter as in the left panel, but with
$10$ extra cliques, each of which contains $10$ randomly selected nodes.
Theses cliques do not carry information about the planted partition, hence 
act as noise to the system. In addition to the non-backtracking matrix,
X-Laplacian, and the adjacency matrix, we put into comparison the results obtained
using other classic and newly proposed matrices, including Bethe Hessian~\cite{saade2014spectral}, Normalized Laplacian~(N. Laplacian) $L_{\textrm{sym}}=I-\tilde A$, and regularized and normalized Laplacian~(R.N. Laplacian)
$L_{A}=\tilde A-\zeta \bf 1\bf 1^T$, with a optimized regularization $\zeta$ (we have scanned the whole range of $\zeta$, and chosen an optimal
one that gives the largest overlap, i.e. fraction of correctly reconstructed labels, in most of cases).
From the figure we see that with the noise added, only X-Laplacian works
down to the original transition (of SBM without cliques). All other 
matrices fail in detecting the community structure with $\epsilon>0.15$.

We have tested other kinds of noisy models, including the
noisy stochastic block model, as proposed in~\cite{Javanmard21032016}. 
Our results show that the X-Laplacian works well (see appendices) while all other spectral methods
do not work at all on this dataset~\cite{Javanmard21032016}.
Moreover, in addition to the classic stochastic block model,
we have extensively evaluated our method on networks
generated by the degree-corrected stochastic block model~\cite{Karrer2011},
and the stochastic block model with extensive triangles.
We basically obtained qualitatively results as in 
Fig.~\ref{fig:community_detection} that the X-Laplacian works as well as
the state-of-the-art spectral methods for the dataset. 
The figures and detailed results can be found at the appendices.

We have also tested real-world networks with an expert division, and found that 
although the expert division
is usually easy to detect by directly using the adjacency matrix,
the X-Laplacian significantly improves the accuracy of detection.
For example on the political blogs network~\cite{Adamic2005}, spectral
clustering using the adjacency matrix gives $83$ mis-classified labels
among totally $1222$ labels, while the X-Laplacian gives only $50$ 
mis-classified labels.
\begin{figure}[h]
   \centering
   {    \includegraphics[width=\hfw\linewidth]{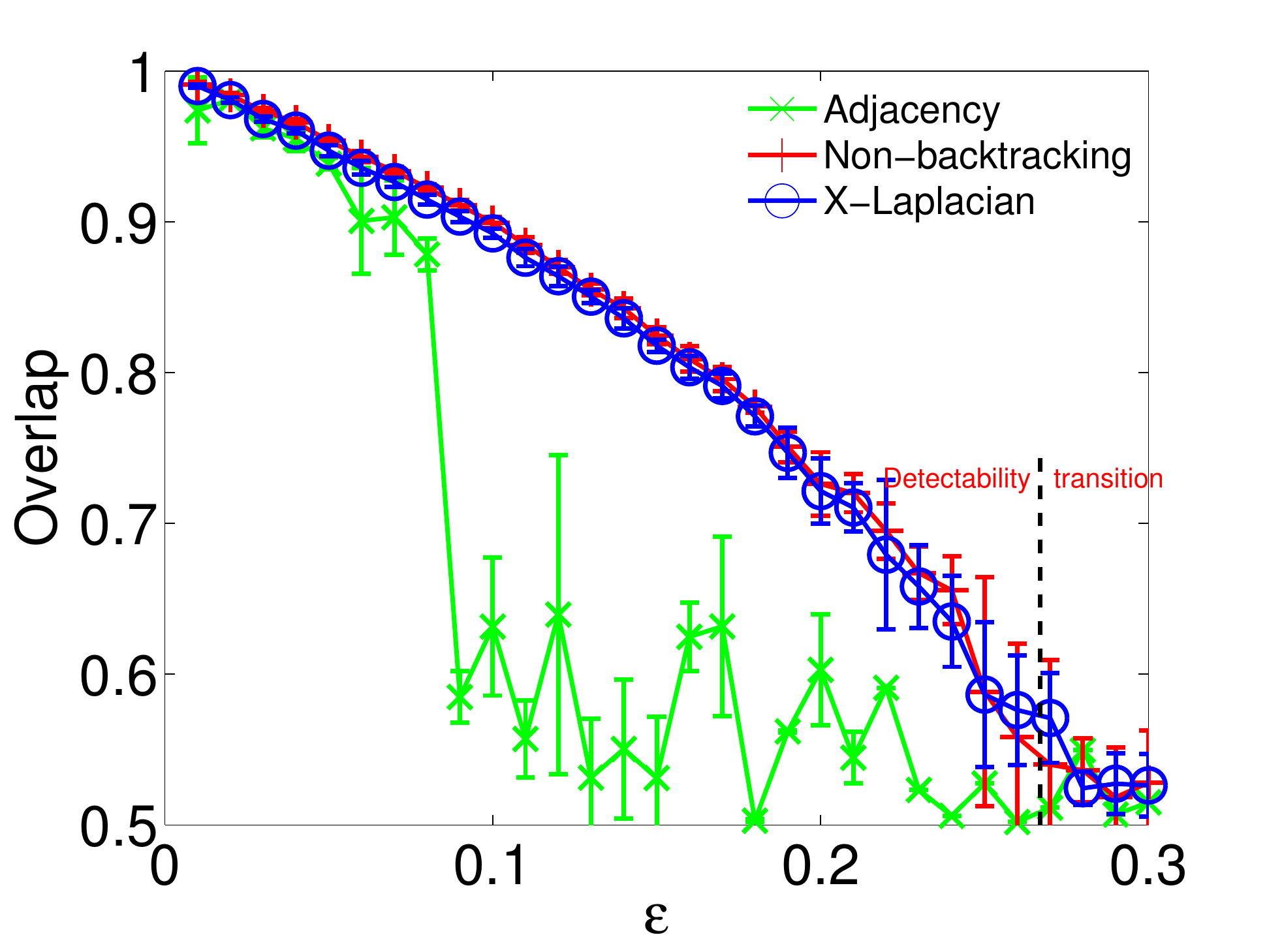} 
	\includegraphics[width=\hfw\linewidth]{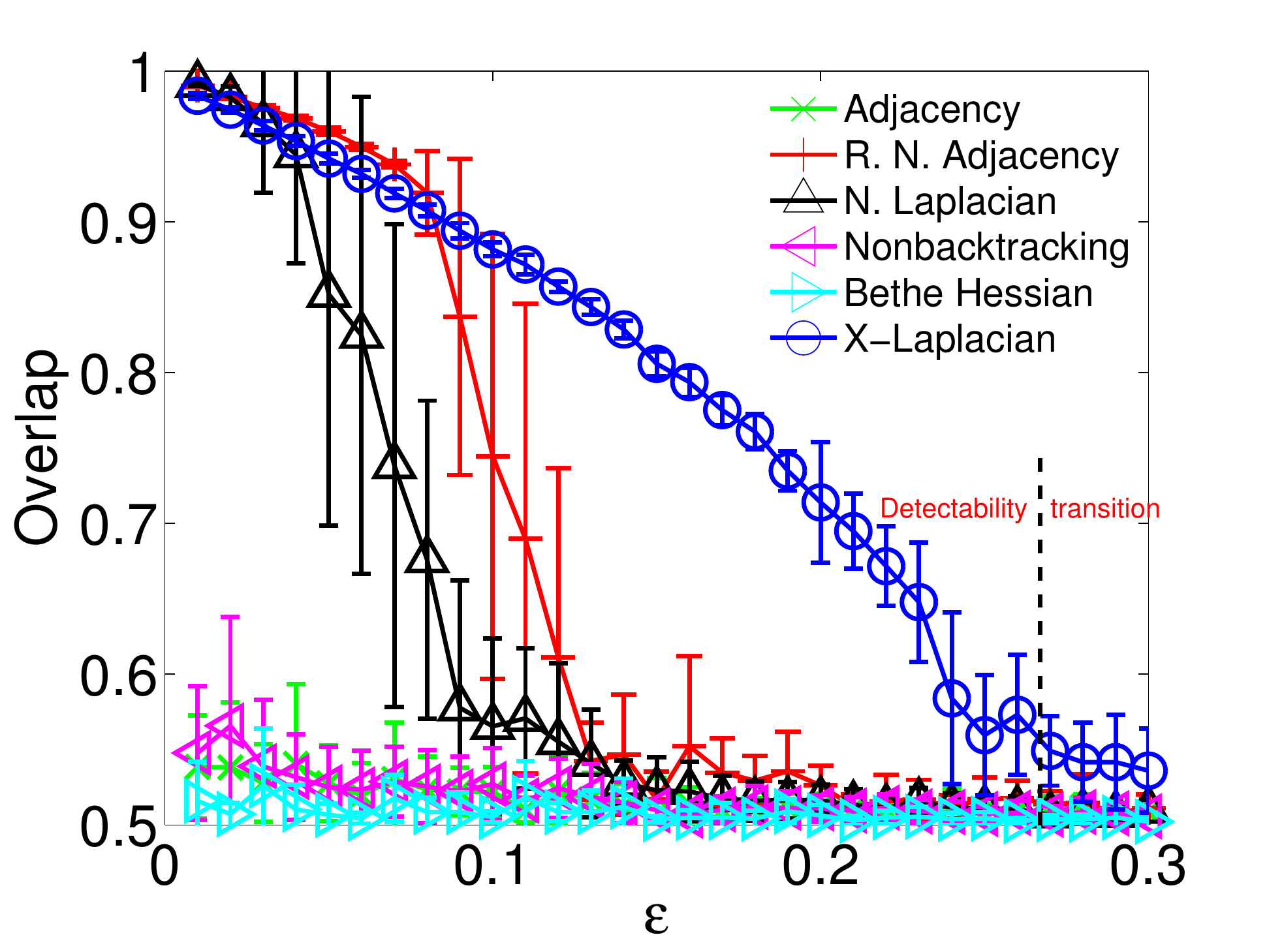} }

	\caption{\label{fig:community_detection}Accuracy of community detection, represented by overlap (fraction of correctly reconstructed labels)
	between inferred partition and the planted partition, for several methods on
	networks generated by the stochastic block model 
	with average degree $c=3$ (\textit{left}) and
with extra $10$ size-$10$ cliques (\textit{right}).
	All networks has $n=10000$ nodes and $q=2$ groups, $\epsilon=\cout/\cin$. 
The black dashed lines denote the theoretical detectability transition. Each data point is averaged over $20$ realizations.
   }
\end{figure}
\subsection{Clustering from sparse pairwise measurements}
Consider the problem of grouping $n$ items into clusters
based on the similarity matrix $S\in\mathbb{R}^{n\times n}$, where
$S_{ij}$ is the pairwise similarity between items $i$ and $j$.
Here we consider not using all pairwise similarities, but only $O(n)$ random samples
of them. In other words, the similarity graph which encodes the 
information of the global clustering structure is sparse, rather than
the complete graph. There are many motivations for choosing such
sparse observations, for example in some cases all measurements are 
simply not available or even can not be stored.

In this section we use the generative model recently proposed in 
\cite{saade2016clustering}, since there is a theoretical limit that
can be used to evaluate algorithms. 
Without loss of generality, we consider
the problem with only $q=2$ clusters. The
model in \cite{saade2016clustering} first assigns items
hidden clusters $\{t_i\} \in \{1, 2\}^n$, then generates similarity
between a randomly sampled pairs of items according 
to probability distribution, $\pin$ and $\pout$, associated with membership of two items.
There is a theoretical limit
$\hat c$ satisfying 
$\frac{1}{\hat c}=\frac{1}{q}\int ds\frac{(\pin(s)-\pout(s))^2}{\pin(s)+(q-1)\pout(s)},$
that with $c<\hat c$ no algorithm could obtain any partial information
of the planted clusters; while with $c>\hat c$ some algorithms, e.g.
spectral clustering using the Bethe Hessian~\cite{saade2016clustering}, achieve partial recovery of the planted
clusters.

Similar to the community detection in sparse graphs, spectral algorithms 
directly using the eigenvectors of a similarity matrix $S$
does not work well, due to the localization of eigenvectors induced
by the sparsity.
To evaluate whether our method, the X-Laplacian, solves the localization problem,
and how it works compared with the Bethe Hessian, in Fig.~\ref{fig:clustering}
we plot the performance (in overlap, the fraction of correctly reconstructed group labels) of three algorithms on the same set of similarity matrices.
For all the datasets there are two groups with distributions $p_{\text{in}}$ and $p_{\text{out}}$
being Gaussian with unit variance and mean $0.75$ and $-0.75$ respectively.
In the left panel of Fig.~\ref{fig:clustering} the topology of pairwise 
entries is random graph, Bethe Hessian 
works down to the theoretical limit, while directly using of the measurement matrix
gives a poor performance. We can also see that X-Laplacian 
has fixed the localization problem of directly using of the measurement matrix, and 
works almost as good as the Bethe-Hessian. We note that
the Bethe Hessian needs to know 
the parameters (i.e. parameters of distributions $\pin$ and $\pout$), while
the X-Laplacian does not use them at all.

In the right panel of Fig.~\ref{fig:clustering}, 
on top of the ER random graph topology, we add some noisy local
structures by randomly selecting $20$ nodes and connecting
neighbors of each selected node to each other.
The weights for the local pairwise were set to $1$, so that the noisy structures
do not contain information about the underlying clustering. We can see that Bethe Hessian is influenced by
noisy local structures and fails to work, while X-Laplacian solves the
localization problems induced by sparsity, and is robust to the noise.
We have also tested other kinds of noise by adding cliques,
or hubs, and obtained similar results (see appendices).
\begin{figure}[h!]
   \centering
    \includegraphics[width=\hfw\linewidth]{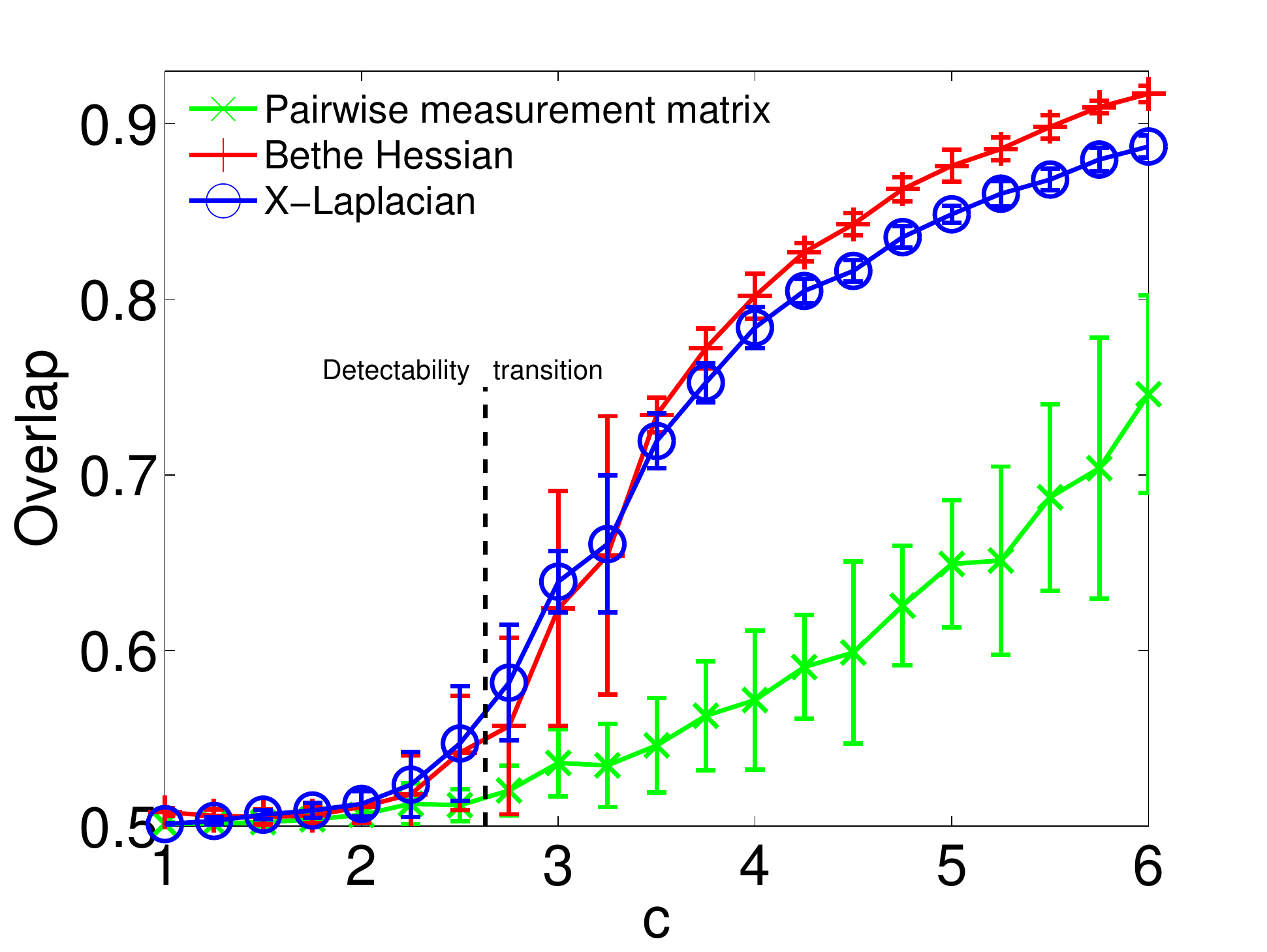} 
    \includegraphics[width=\hfw\linewidth]{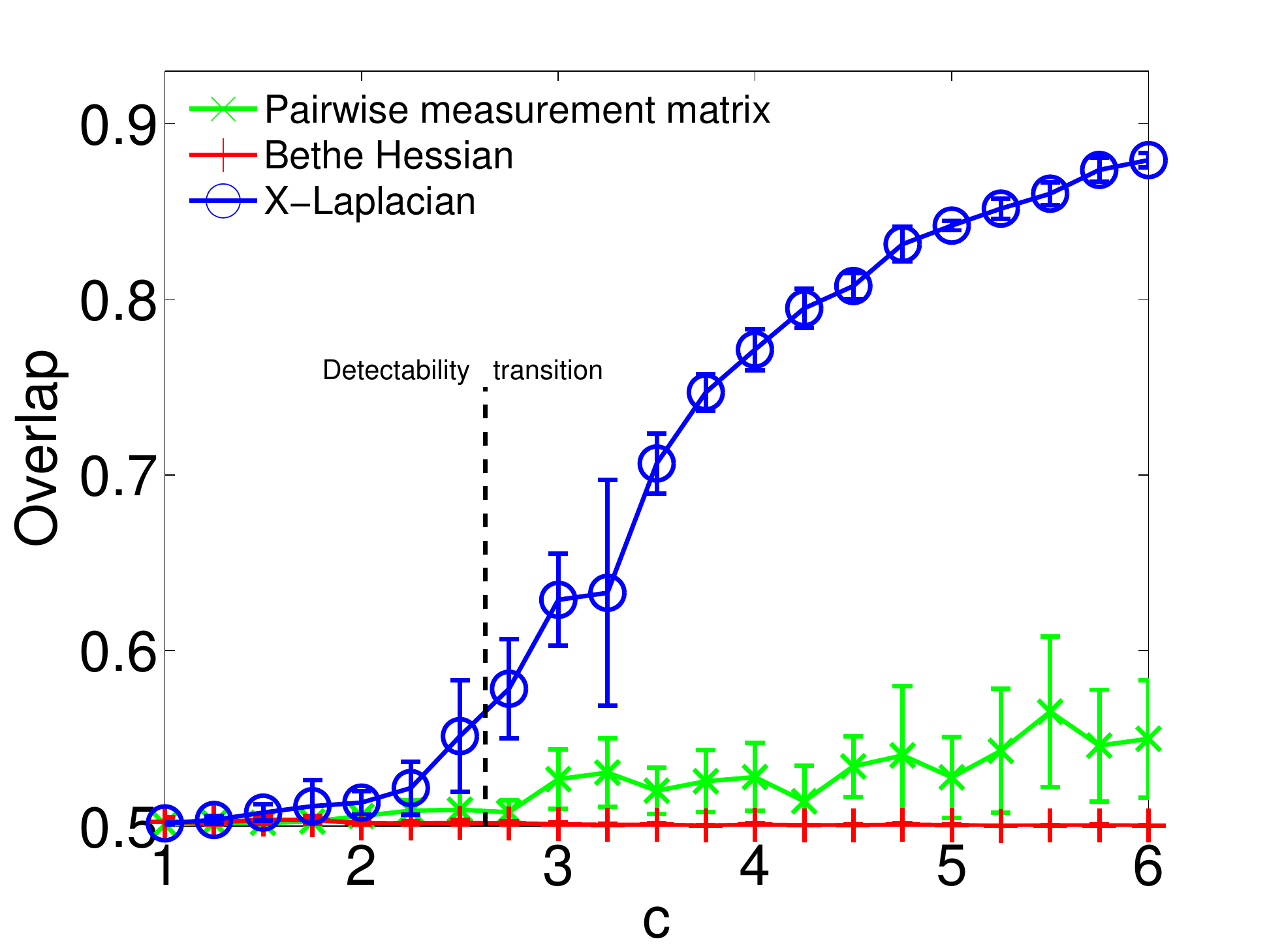} 
	\caption{Spectral clustering using sparse pairwise measurements. 
	The X-axis denotes the average number of pairwise measurements 
	per data point, and 
	the Y-axis is the fraction of correctly reconstructed labels, maximized
	over permutations.
	The model used to generate pairwise measurements is
	proposed in~\cite{saade2016clustering}, see text for detailed descriptions.
	In the left panel, the topologies of the pairwise measurements are random graphs. In the right panel 
in addition to the random graph topology there are 
$20$ randomly selected nodes with all their neighbors connected.
	Each point in the figure is averaged over $20$ realizations of
	size $10^4$.
   \label{fig:clustering}
	}
\end{figure}

\subsection{Rank estimation and Matrix Completion}
The last problem we consider in this paper for evaluating the X-Laplacian is
completion of a low rank  matrix from few entries. This problem
has many applications including the famous collaborative filtering.
A problem that is closely related to it is the
rank estimation from revealed entries.
Indeed estimating rank of the matrix is usually the first step before 
actually doing the matrix completion.
The problem is defined as follows: let $A^{\textrm{true}}=UV^T$,
where $U\in\mathbb{R}^{n\times r}$ and $V\in\mathbb{R}^{m\times r}$ are
chosen uniformly at random and $r\ll \sqrt{nm}$ is the ground-true rank. Only few, say $c\sqrt{mn}$, entries of matrix $A^{\mathrm{true}}$
are revealed. That is we are given a matrix $A\in\mathbb{R}^{n\times m}$
who contains only subset of $A^{\textrm{true}}$, 
with other elements being zero. 
Many algorithms have been proposed for matrix completion, including nuclear norm minimization~\cite{candes2009exact} and methods based on the singular value decomposition~\cite{cai2010singular} etc.
Trimming which sets to zero all rows and columns with a large
revealed entries, is usually introduced to control the localizations of singular vectors and to estimate the rank using
the gap of singular values~\cite{keshavan2009matrix}.
Analogous to the community detection problem, trimming is not 
supposed to work optimally when matrix $A$ is sparse.
Indeed in ~\cite{saade2015matrix} authors reported that 
their approach based on the Bethe Hessian outperforms trimming+SVD 
when the 
topology of revealed entries is a sparse random graph.
Moreover, authors in ~\cite{saade2015matrix} show that the 
number of negative eigenvalues of the Bethe Hessian gives a more accurate
estimate of the rank of $A$ than that based on trimming+SVD.

\begin{figure}[h!]
   \centering
    \includegraphics[width=\hfw\linewidth]{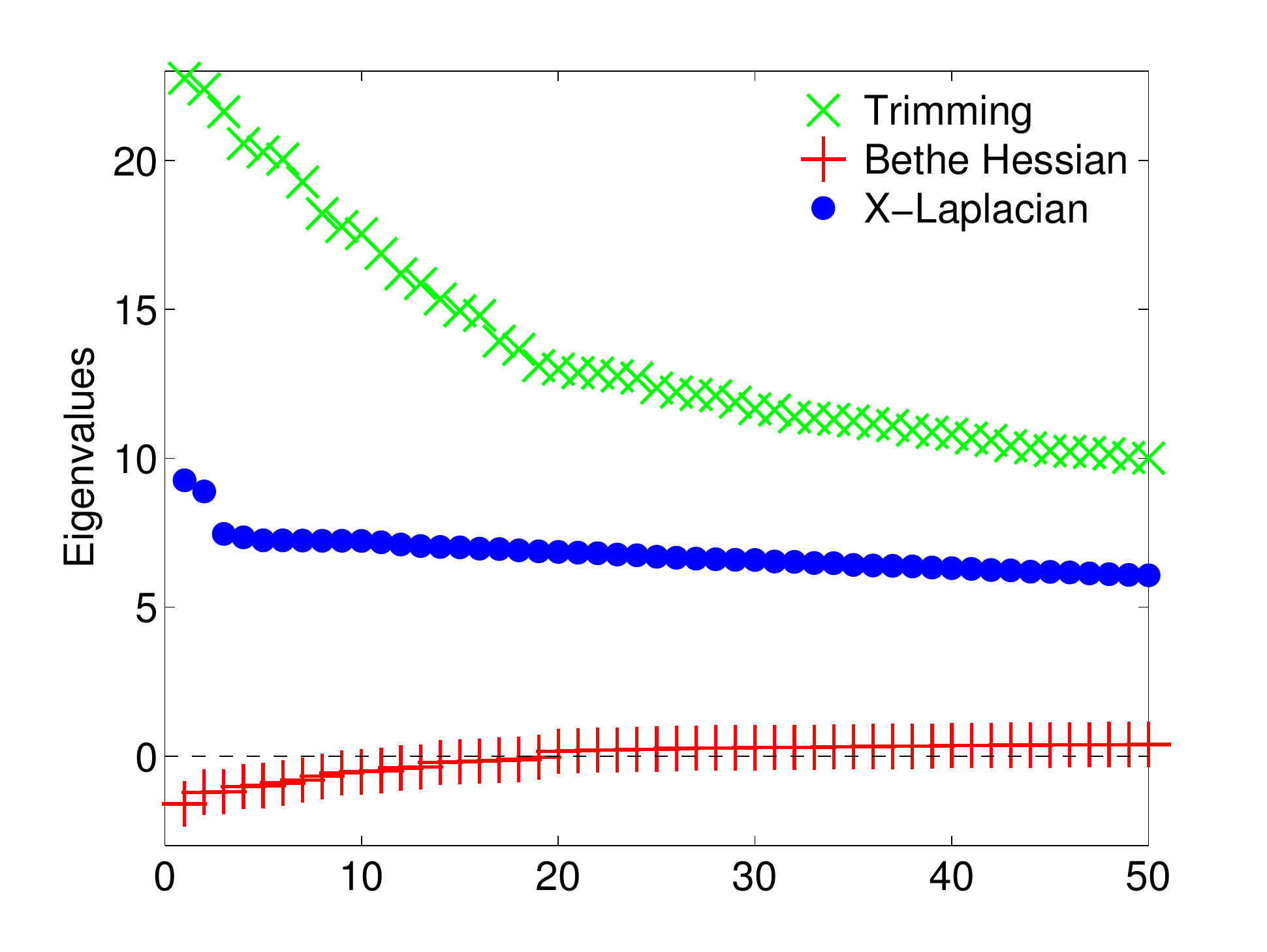} 
	\includegraphics[width=\hfw\linewidth]{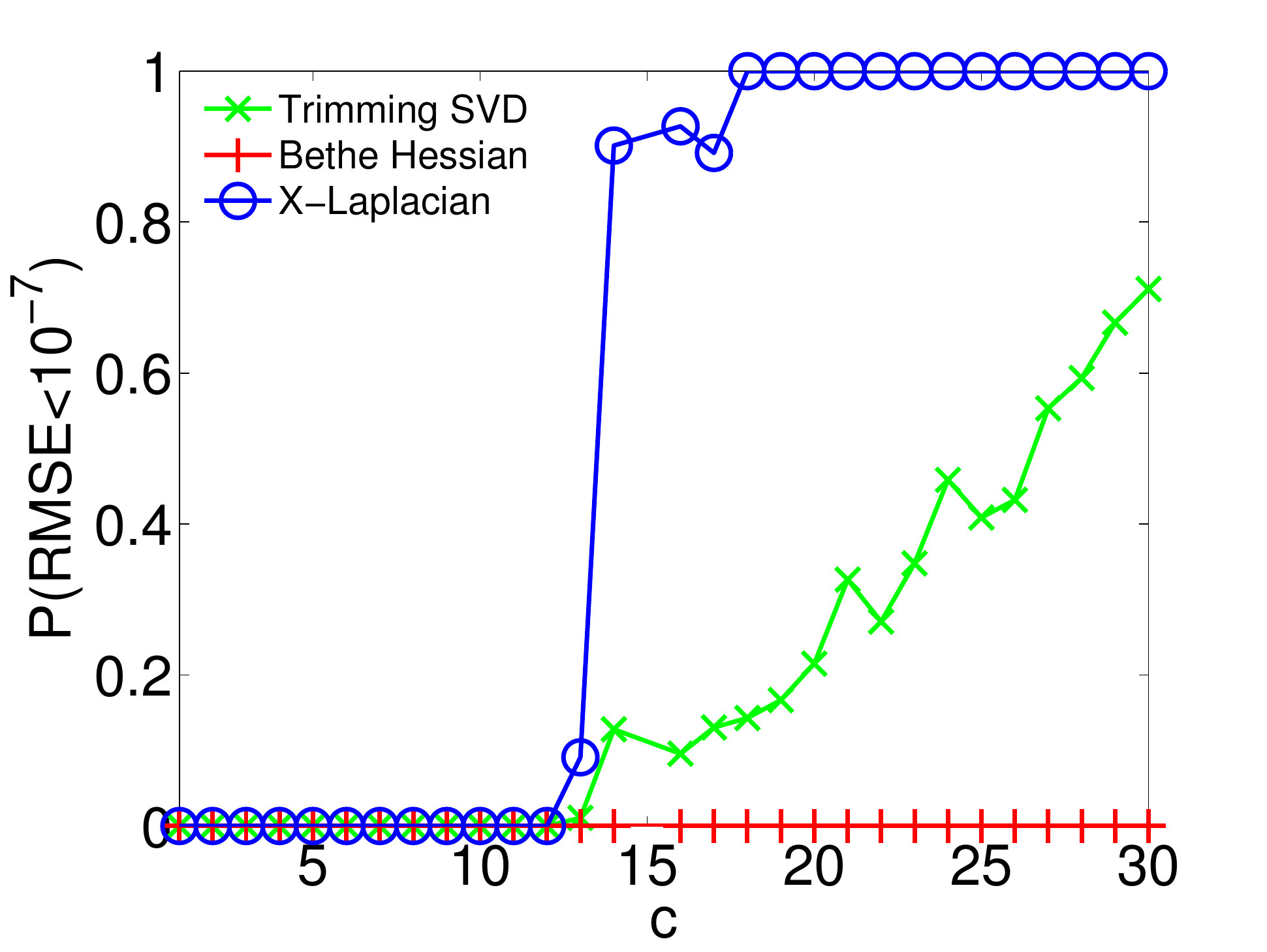} 
	\caption{\textit{(Left:) }Singular values of sparse data matrix with trimming, eigenvalues of the Bethe
	Hessian and X-Laplacian. The data matrix is the outer product of two vectors of
	size $1000$. Their entries are Gaussian random variables with mean zero and unit variance, so the rank of the original matrix is $2$.
	The topology of revealed observations are random graphs with average
	degree $c=8$ plus $10$ random cliques of size $20$. 
	\textit{(Right:)} Fraction of samples that RMSE
	is smaller than $10^{-7}$, among $100$ samples of rank-$3$ data matrix $UV^T$ of size
	$1000\times 1000$, with the entries of $U$ and $V$ drawn from a Gaussian
	distribution of mean $0$ and unit variance. The topology of revealed entries is the random graph with 
	varying average degree $c$ plus $10$ size-20 cliques. 
   \label{fig:rank}
	}
\end{figure}

However, we see that if the topology is not locally-tree-like but with 
some noise, for example 
with some additional cliques, both trimming of the data matrix and 
Bethe Hessian perform much worse, reporting a wrong rank, and giving a 
large reconstruction error, as illustrated in Fig.~\ref{fig:rank}.
In the left panel of the figure we plot the eigenvalues of
the Bethe Hessian, and singular values of trimmed matrix $A$ with
true rank $r^{\textrm{true}}=2$.
We can see that both of them are continuously distributed: there is no 
clear gap in singular values of trimmed $A$, and Bethe Hessian has 
lots of negative eigenvalues.
In this case since matrix $A$ could be a non-squared matrix, we need to
define the X-Laplacian as $L_X=\begin{pmatrix}
	0 & A \\A&0\end{pmatrix}-X$.
The eigenvalues of $L_X$ are also plotted in Fig.~\ref{fig:rank} where
one can see clearly that
there is a gap between the second largest eigenvalue and the third one.
Thus the correct rank can be estimated using the value minimizing consecutive eigenvalues, 
as suggested in \cite{keshavan2009matrix}.

After estimating the rank of the matrix, matrix completion 
is done by using a local optimization algorithm \cite{nlopt}
starting from initial matrices, that 
obtained using first $r$ singular vectors of trimming+SVD,
first $r$ eigenvectors of Bethe Hessian and X-Laplacian with estimated rank $r$ respectively. The results are shown in Fig.~\ref{fig:rank} right where
we plot the probability that 
obtained root mean square error (RMSE) is smaller than $10^{-7}$ as a function of average number 
of revealed entries per row $c$, 
for the ER random-graph topology plus noise represented by 
several cliques. We can see that
X-Laplacian outperforms Bethe Hessian and Trimming+SVD with $c\geq 13$. 
Moreover, when $c\geq 18$, for all instances, only X-Laplacian 
gives an accurate completion for all instances.

\section{Conclusion and discussion}
We have presented the X-Laplacian, a general approach
for detecting latent global structure in a given data matrix. It is 
completely a data-driven approach that learns different forms of 
regularization
for different data, to solve the problem of localization of
eigenvectors or singular vectors.
The mechanics for de-localizing of eigenvectors during learning of 
regularizations has been illustrated using the matrix perturbation analysis.
We have validated our method using extensive numerical 
experiments, and shown that it 
outperforms state-of-the-art algorithms on various inference problems
in the sparse regime and with noise.

In this paper we discuss the X-Laplacian using directly the data matrix
$A$, but the this is not the only choice.
Actually we have tested approaches using various variants of A,
such as $\tilde A$, and found they work as well.
We have also tried learning regularizations for the Bethe Hessian, and 
found it succeeds in repairing Bethe Hessian when Bethe Hessian has localization problem.
These indicate that our scheme of regularization-learning is a general
spectral approach for hard inference problems.
\small
\bibliographystyle{abbrv}

\section{Appendix}
\subsection{Perturbation analysis}
After applying the perturbation, 
We anticipate that an eigenvalue of
$L_X$ changes from $\lambda_i$ to $\lambda_i + \hat \lambda_i$, and an eigenvector changes
from $u_i$ to $u_i+\hat u_i$. If we assume that matrix $L_X$ is not ill-conditioned, and the
first few eigenvectors that we care about are distinct, then we have
$$(L_X+\hat L)(u_i+\hat u_i)=(\lambda_i+\hat \lambda_i)(u_i+\hat u_i).$$
By making use of
$$L_Xu_i=\lambda_i u_i,$$
and keeping only first order terms, we have
\begin{equation}\label{eq:pert1}\hat Lu_i+L_X\hat u_i=\lambda_i \hat u_i+\hat \lambda_i u_i.\end{equation}
Since $L_X$ is a real symmetric matrix, we can represent $\hat u_i$ as a weighted sum of 
eigenvectors of $L_X$, as
\begin{equation}\label{eq:pert2}\hat u_i=\sum_{j=1}^n\omega_ju_j,\end{equation}
where $\omega_j$ is the coefficient and $u_j$ is j'th eigenvector of $L_X$.
Insert last equation into Eq.~\eqref{eq:pert1}, we have
\begin{equation}\label{eq:pert3}\hat Lu_i+L_X\sum_{j=1}^n\omega_ju_j=\lambda_i \sum_{j=1}^n\omega_ju_j+\hat \lambda_i u_i,\end{equation}
which evaluates to
\begin{equation}\label{eq:pert4}\hat Lu_i+\sum_{j=1}^n\omega_j\lambda_ju_j=\lambda_i \sum_{j=1}^n\omega_ju_j+\hat \lambda_i u_i.\end{equation}
Multiplying $u_i^T$ to both sides of last equation results to
\begin{equation}\label{eq:pert5}u_i^T\hat Lu_i+\sum_{j=1}^n\omega_j\lambda_ju_i^Tu_j=\lambda_i \sum_{j=1}^nu_i^T\omega_ju_j+\hat \lambda_i u_i^Tu_i.\end{equation}
Notice that in the last equation $u^T_iu_i=1$ and the second term in the left hand side and the first term
in the right hand side cancel each other, thus we have
\begin{equation}\label{eq:pert6}
	\hat \lambda_i=u_i^T\hat Lu_i.
\end{equation}
In our algorithm, $\hat L$ is a diagonal matrix with entries $\hat L_{ii}=-\eta v_{i}^2$ where $v_{i}$ denotes the $i$'th element of 
the selected eigenvector $v$ who has the largest inverse participation ratio.
Thus the shift of an eigenvalue $\lambda_j$ associated with eigenvector $u_j$ (which is different from $v$) is then
\begin{equation}
	\hat \lambda_j=-\eta \sum_{i=1}^nv_{i}^2u_{ji}^2.
\end{equation}
For the selected vector $v$, the change of its eigenvalue is
\begin{equation}
	\hat \lambda_v=-\eta \sum_{i=1}^nv_{i}^4=-\eta I(v).
\end{equation}
That is, the amount of decreasing of eigenvalue associated with 
the selected vector is proportional to its inverse participation
ratio.

In addition to the shift of eigenvalues, we can also derive
the change of eigenvectors after perturbation.
Multiplying transpose of an eigenvector $u_j$ to both sides of Eq.\eqref{eq:pert4} 
results to 
\begin{equation}\label{eq:pert7}
	u_j^T\hat Lu_i+\sum_{k=1}^n\omega_k\lambda_ku_j^Tu_k=\lambda_i \sum_{k=1}^nu_j^T\omega_ku_k,
\end{equation}
which evaluates to
\begin{equation}\label{eq:pert8}
	u_j^T\hat Lu_i+\omega_j\lambda_j=\lambda_i\omega_j, 
\end{equation}
where we can find that 
\begin{equation}\label{eq:pert9}
\omega_j=\frac{u_j^T\hat Lu_i}{\lambda_i-\lambda_j}.
\end{equation}
Given that the perturbation is $\hat L_{ii}=-\eta v_i^2$, we have an expression for the change of an eigenvector
\begin{align}\label{eq:pert10}
	\hat {u_i}&=\sum_{j\neq i}\frac{u_j^T\hat Lu_i}{\lambda_i-\lambda_j}u_j\nonumber\\
	&=-\eta\sum_{j\neq i}\frac{\sum_ku_{jk}v_k^2u_{ik}}{\lambda_i-\lambda_j}u_j.
\end{align}
Notice that the inverse participate ratio of the new vector $u_i+\hat u_i$ is
\begin{align}
	I(u_i+\hat u_i)&=\sum_{l=1}^n(u_{il}+\hat u_{il})^4,\nonumber\\
\end{align}
Expand above equation to the first order of $\hat u_{il}$, we have
\begin{align}
	I(u_i+\hat u_i)&\approx I(u_i)+4\sum_{l=1}^nu_{il}^3\hat u_{il}\nonumber\\
	&=I(u_i)-4\eta\sum_{l=1}^nu_{il}^3\sum_{j\neq i}\frac{\sum_ku_{jk}v_k^2u_{ik}}{\lambda_i-\lambda_j}u_{jl}\nonumber\\
	&=I(u_i)-4\eta\sum_{l=1}^n\sum_{j\neq i}\frac{u^2_{jl}v_l^2u^4_{il}}{\lambda_i-\lambda_j}
	-4\eta\sum_{l=1}^n\sum_{j\neq i}\sum_{k\neq l}\frac{u_{il}^3v_k^2u_{jk}u_{ik}u_{jl}}{\lambda_i-\lambda_j}
	\label{eq:signal}
\end{align}

\subsection{Detailed process of learning a regularization}
\begin{figure}[h]
   \centering
   \label{fig:process1}
    \includegraphics[width=0.31\linewidth]{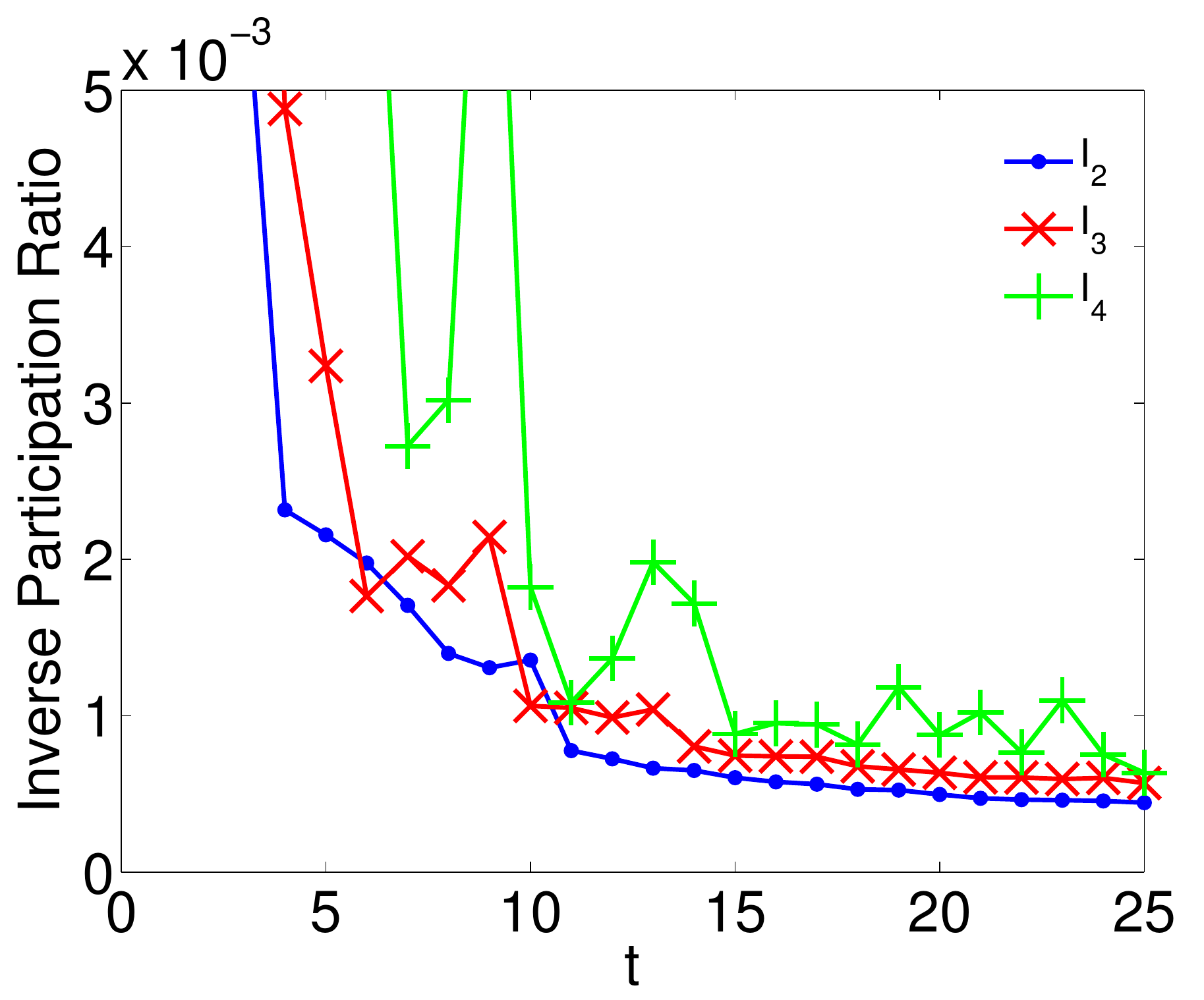} 
    \includegraphics[width=\tfw\linewidth]{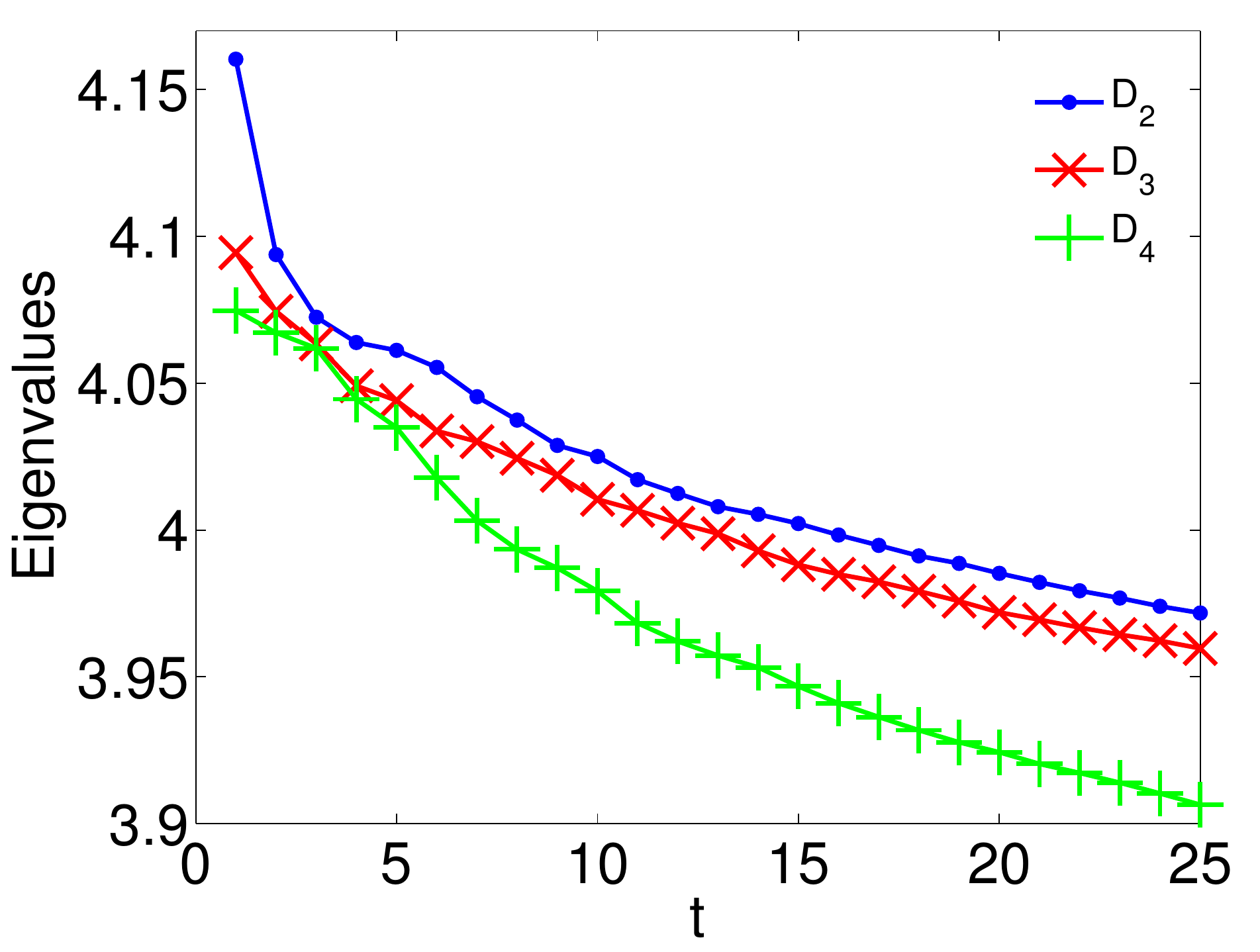} 
    \includegraphics[width=\tfw\linewidth]{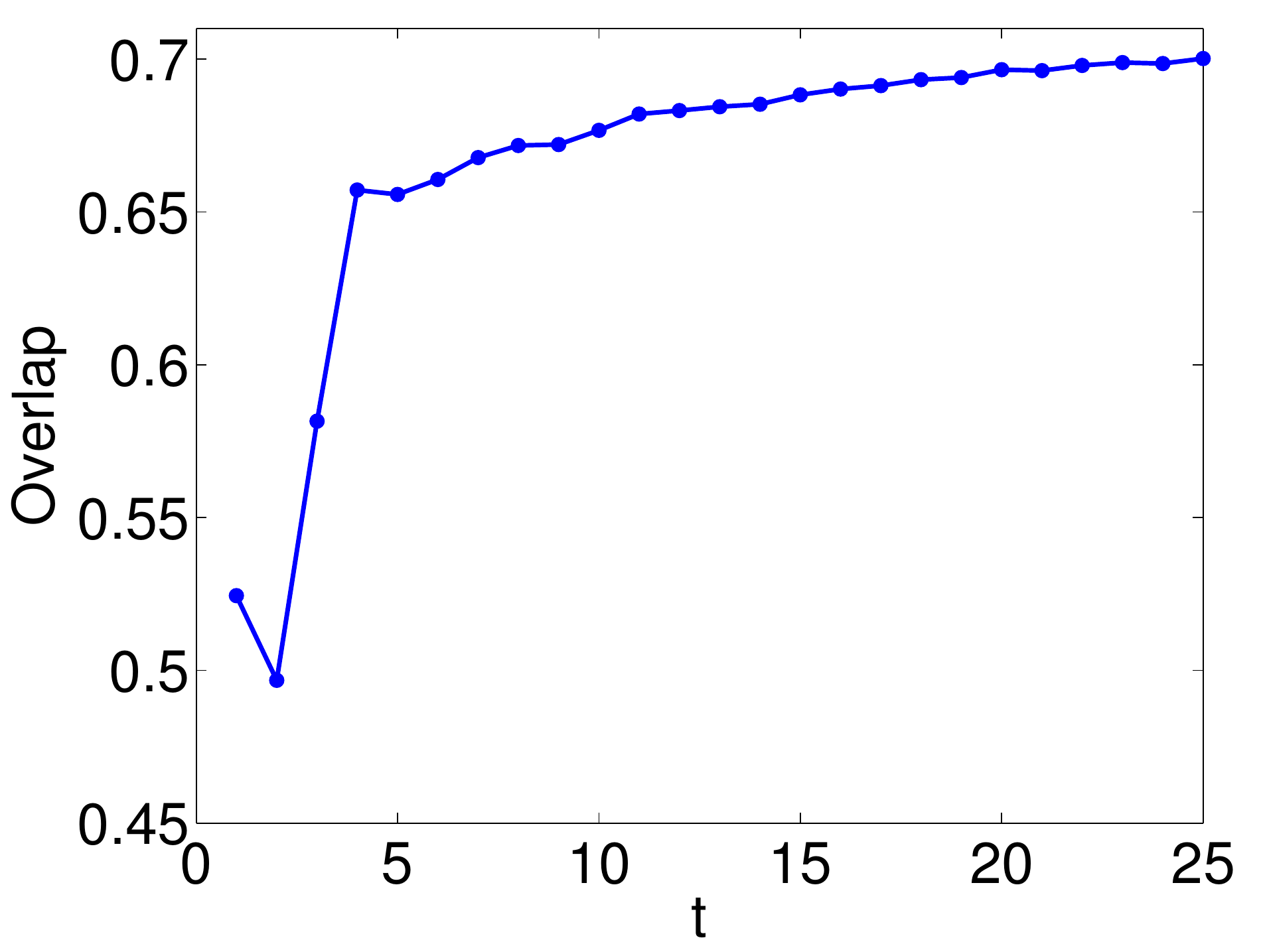} 
	\caption{Inverse Participation Ratio of first three eigenvectors ($I_1, I_2, I_3$, overlap (the fraction of correctly reconstructed labels) and first three eigenvalues ($D_1, D_2, D_3$) as a function of learning steps
	$t$ for a network generated by the stochastic block model with $n=42000$, $q=3$ groups, average degree $c=3$, $\epsilon=0.08$ and
learning rate $\eta=1$. The overlap is the fraction of successfully 
reconstructed labels, maximized over group permutations.}
\end{figure}

In Fig.~\ref{fig:process1} we plot the evolution of eigenvalues, overlap
and the Inverse Participation Ratio (IPR) for the second, third and forth
eigenvectors
during learning of the X-Laplacian for a network generated by
the stochastic block model.
The network has a community structure with $3$ groups, however the first
three eigenvectors of the adjacency matrix are localized (see left panel at
$t=0$) and 
do not reveal the underlying community structure (see the right panel 
at $t$ small.
We can also see from the left panel that
the IPR of them are decreasing as $t$ increases during learning.
From the middle panel of the figure, we see that all the $3$ eigenvalues
are decreasing, while the spectral gap $D_3-D_4$ is increasing during 
learning. 
It is interesting to see that at $t=4$, there is a exchange of positions
of the third eigenvector and the forth eigenvector. This gives a bump
of the IPR, as well as an increase of accuracy of detection (characterized
by overlap) at $t=4$.

\subsection{Additional numerical evaluations on community detections}
Here we compare the performance of the X-Laplacian with other state-of-art
spectral algorithms on variants of the stochastic block model, namely
the degree-corrected stochastic block model~\cite{Karrer2011} and 
the triangular stochastic block model~\cite{tsbm} which is the 
stochastic block model with triangles.
It is known that in the stochastic block model, there is a 
detectability transition at 
$$\epsilon^*=(\sqrt{\hat c}-1)/(\sqrt{\hat c}-1+q),$$ 
where $\hat c$ is the excess average degree
$$\hat c=\frac{\brc{k^2}}{\brc{k}}-1,$$
and the spectral clustering algorithm based on the non-backtracking
matrix achieves this threshold.
In the left panel of~Fig.\ref{fig:community_detection} we compare 
the performance (evaluated using the overlap, fraction of correctly 
reconstructed labels) of spectral algorithms using the adjacency matrix,
the non-backtracking matrix and the X-Laplacian on networks generated
by the degree corrected stochastic block model with a power-law
degree distribution with exponent $-2.5$. As the figure shows, our
approach works even better than the algorithm using the non-backtracking
matrix, this is because when the networks size ($10^4$) is not large 
enough, the long tails of degree distribution creates short loops in
the network, downgrading the performance of the algorithm using the
non-backtracking matrix which is supposed to work optimally in the 
locally-tree like networks.

For the triangular stochastic block model, due to the presence of 
triangles, the non-backtracking matrix suffers
from short loops and does not work well. In this case the generalized non-backtracking
matrix, which runs on a factor graph with both edges and triangles treated as 
function nodes, works down to the transition~\cite{tsbm}.
In the right panel of Fig.~\ref{fig:community_detection} we 
compare the performance of the spectral algorithm using 
the adjacency matrix, generalized non-backtracking matrix and the X-Laplacian
, and we can see that X-Laplacian works as well as the generalized 
non-backtracking matrix and down to the transition.

\begin{figure}[h!]
   \centering
	\includegraphics[width=\fw\columnwidth]{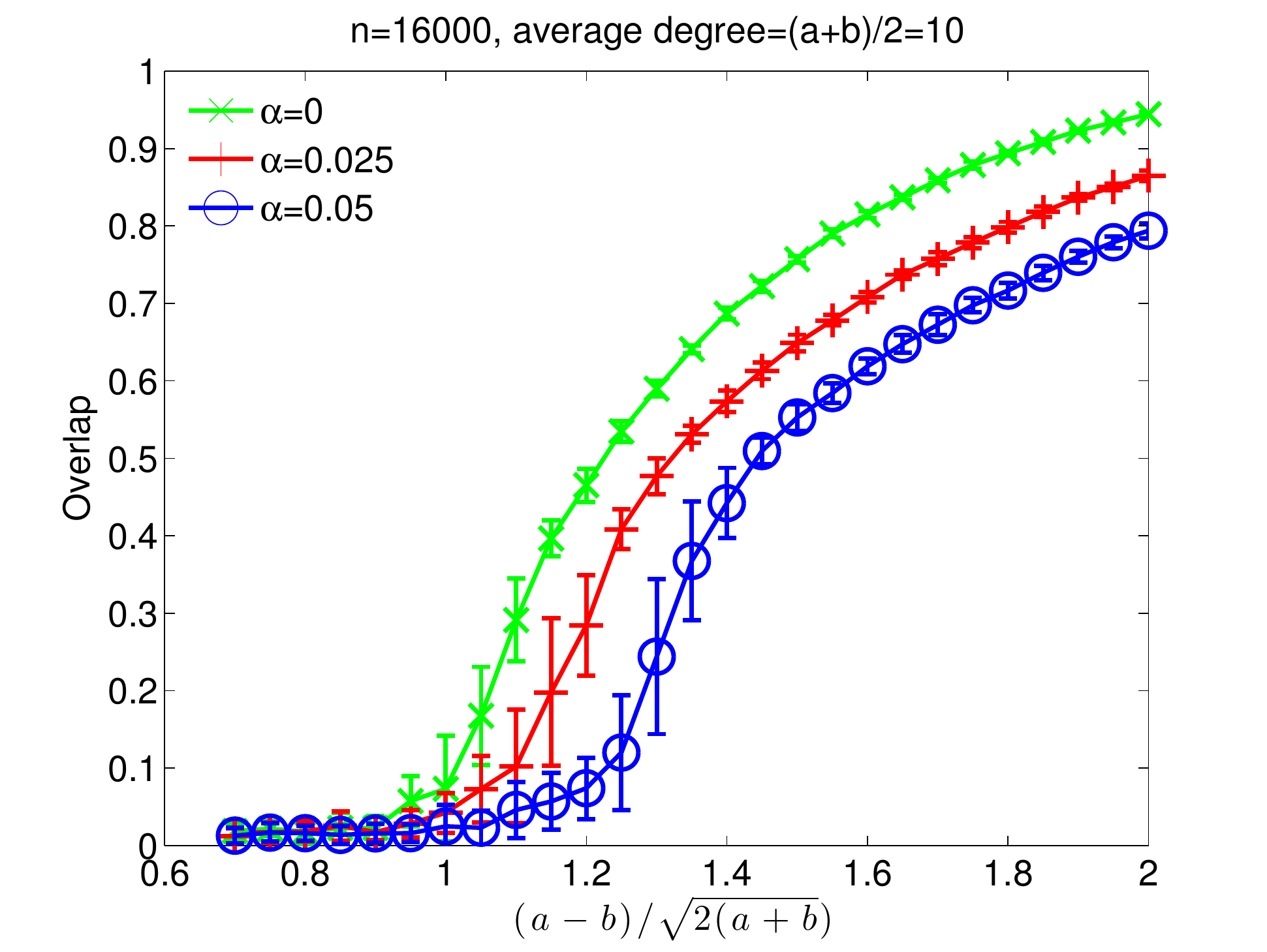} 
	\caption{
	\label{fig:sbm_noise}
	X-Laplacian using $\tilde A$ on networks generated by the perturbed
	stochastic block model \cite{Javanmard21032016}, with exactly the same network size and 
	parameters as in Fig.$5$ of appendices in \cite{Javanmard21032016}.
 $a$ and $b$ denote average degree connecting nodes in the
same group and different groups respectively, and $\alpha$ is the fraction
of selected noisy nodes.
	Each point
	is averaged over $20$ instances.
	}
\end{figure}

It has been reported in ~\cite{Javanmard21032016} that on the perturbed
stochastic block model, spectral algorithms including the one using Bethe Hessian do
fail in detecting the community structures, while other method, e.g. 
semi-definite programming, works well.
In the perturbed stochastic block model, after a network is generated
by the stochastic block model, neighbors of some randomly selected nodes are 
connected to each other acting as noise to the underlying community 
structure, 
In Fig.~\ref{fig:sbm_noise} we numerically examined the performance of
X-Laplacian using $\tilde A$, i.e. $L_X=\tilde A+X$, on 
networks generated by perturbed stochastic block model~
\cite{Javanmard21032016}, with exactly the same network size and 
parameters as in \cite{Javanmard21032016} (see Fig. $5$ in their appendices),
with parameter $a$ and $b$ denoting expected number of edges per node 
connecting nodes in the same group and in different groups, respectively.
By comparing Fig.~\ref{fig:sbm_noise} with Fig. $5$ in appendices of \cite{Javanmard21032016} 
we can see from X-Laplacian works similarly to the semi-definite programming
while Bethe-Hessian based method does not work at all.

\begin{figure}[h]
   \centering
	\includegraphics[width=\hfw\columnwidth]{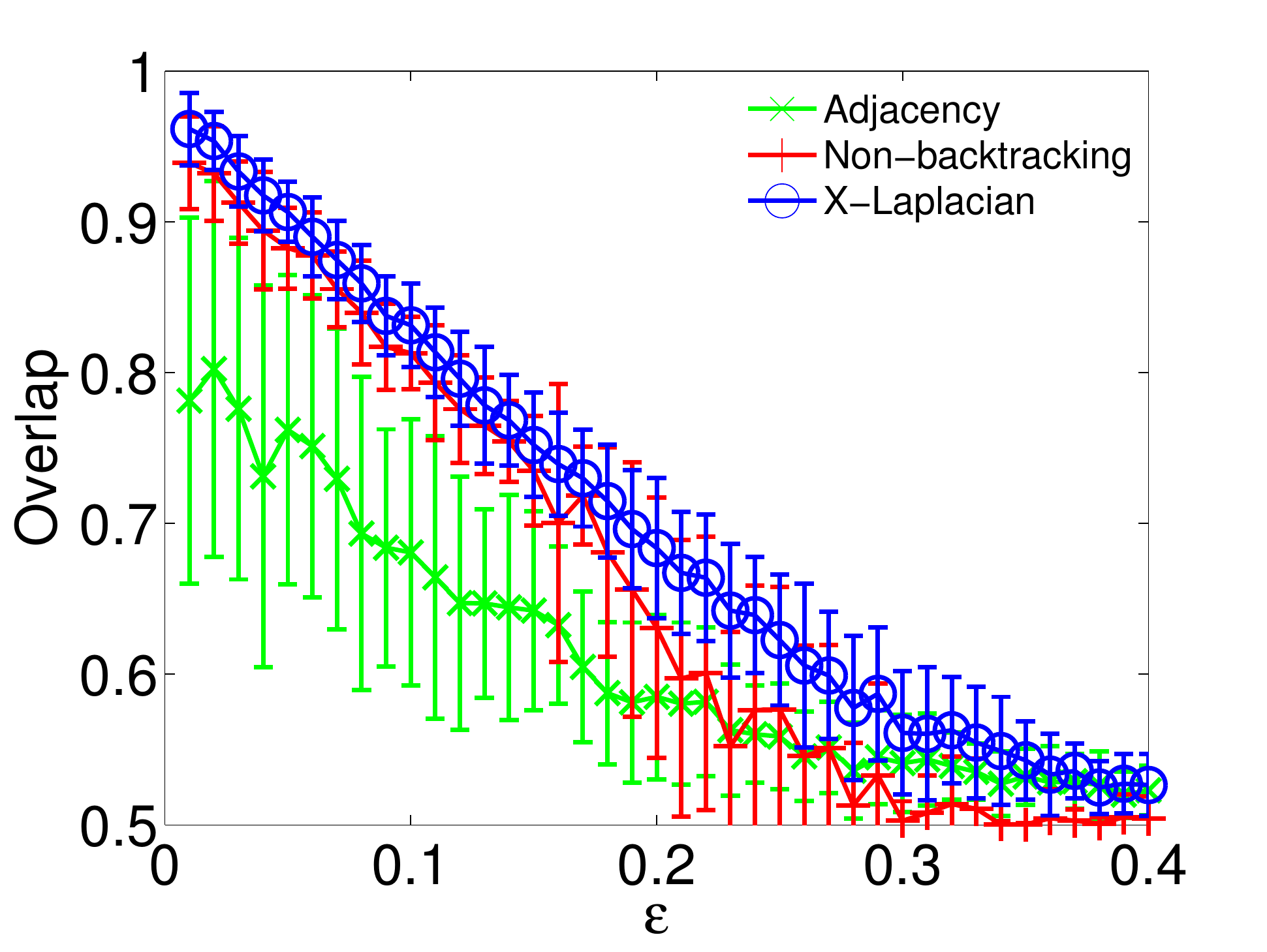} 
    \includegraphics[width=\hfw\columnwidth]{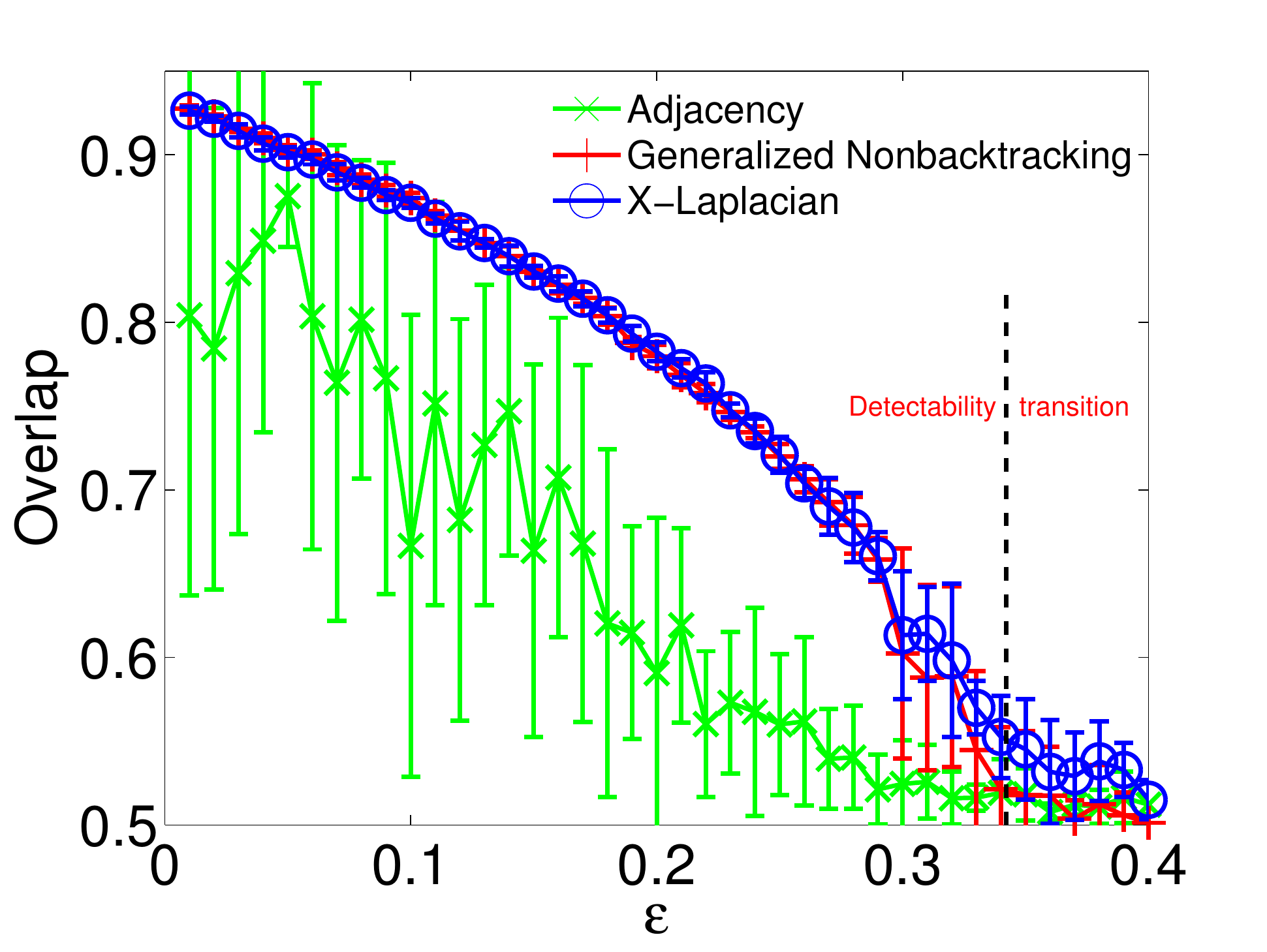} 
	\caption{Accuracy of community detection, represented by overlap
	between inferred partition and the planted partition, for several method on
	networks generated by the 
degree corrected stochastic block model, 
	and a power-law degree distribution with 
	exponent $-2.5$ (\textit{left}); 
	triangular stochastic block model (\textit{right}) with average degree
	$c=3$ and $\rho=0.5$ which means half of edges belong to triangles
	rather than single edges~[unpublished].
	All networks has $n=10000$ nodes and $q=2$ groups. In $X$-axis, 
	$\epsilon=\cout/\cin$ controls the hardness of the problem. 
	Each data point is averaged over $20$ realizations.
   \label{fig:community_detection}
	}
\end{figure}

\subsection{Additional numerical evaluations on spectral clustering using
pairwise similarity measurements}
In this section we compare the performance of spectral algorithms using
the data matrix, the Bethe Hessian and the X-Laplacian, on the model recently proposed in~\cite{saade2015matrix}, which generates 
pairwise measurements between two groups of nodes from different 
probability distributions.
Two distributions $p_{\text{in}}$ and $p_{\text{out}}$ are chosen to be
Gaussian with unit variance and mean $0.75$ and $-0.75$ respectively.
On top of the network we add two different kinds of noise, i.e. cliques
and hubs to the random graph topology. And from figures we can see 
that the results are qualitatively similar to right panel of Fig. $4$ in the main text
where X-Laplacian outperforms both Bethe Hessian and X-Laplacian in
reconstructing the planted partition.

\begin{figure}[h!]
   \centering
	\includegraphics[width=\hfw\columnwidth]{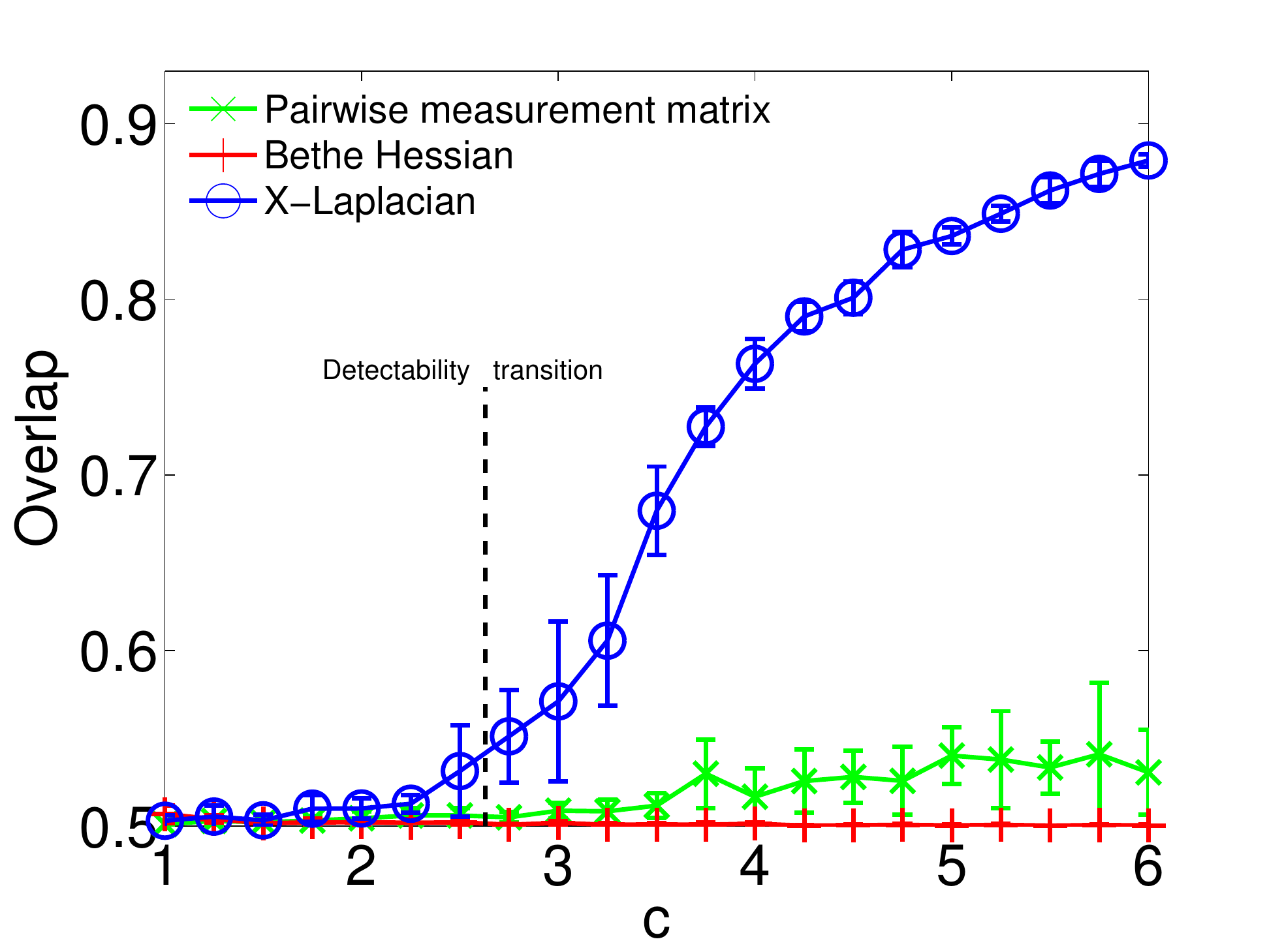} 
	\includegraphics[width=\hfw\columnwidth]{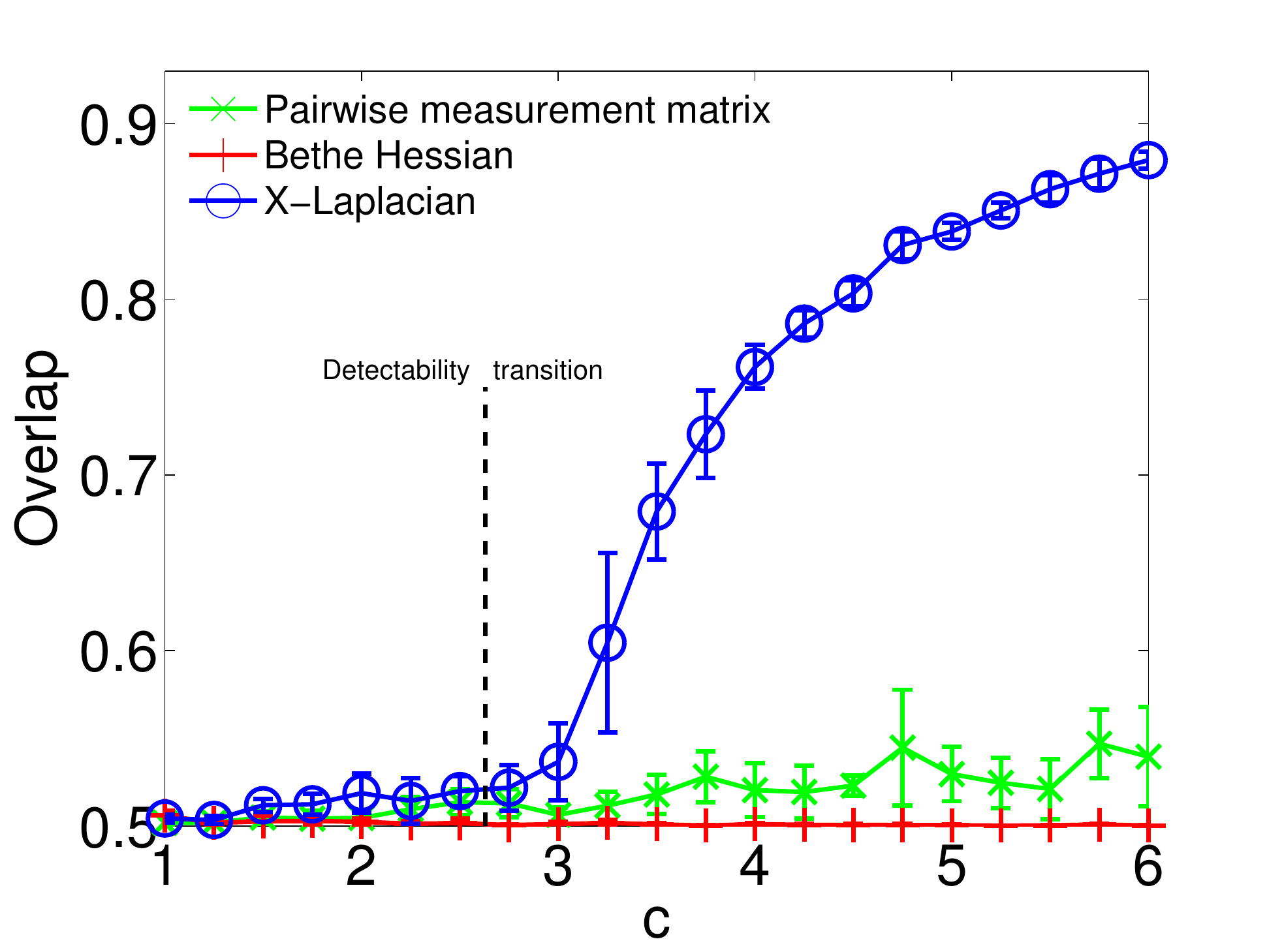} 
	\caption{Spectral clustering using sparse pairwise measurements, using
	model proposed in~\cite{saade2015matrix}.
	The overlap in
	Y-axis is the fraction of correctly reconstructed labels, X-axis denotes
	the average number of pairwise measurements per data point.
		(\textit{Left}): the topologies are random
	graphs with average degree $c$ together with $10$ size-$20$ cliques. 
	(\textit{Right}): panel the topologies are random
	graphs with average degree $c$ together with $10$ hubs whose degrees are $50$.
	Each point in the figure is averaged over $20$ realizations of data set 
	of size $10^4$.
   \label{fig:clustering}
	}
\end{figure}

\end{document}